\documentclass{article}

   \usepackage[final,nonatbib]{neurips_2024}

\usepackage[utf8]{inputenc} 
\usepackage[T1]{fontenc}    
\usepackage{hyperref}       
\usepackage{url}            
\usepackage{booktabs}       
\usepackage{amsfonts}       
\usepackage{nicefrac}       
\usepackage{microtype}      
\usepackage{xcolor}         

\usepackage{graphicx}
\usepackage{caption}

\usepackage{amsmath}
\usepackage{amssymb}

\usepackage{authblk}
\usepackage[misc]{ifsym}

\newcommand{\beginsupplement}{%
        \setcounter{table}{0}
        \renewcommand{\thetable}{S\arabic{table}}%
        \setcounter{figure}{0}
        \renewcommand{\thefigure}{S\arabic{figure}}
        }

\title{Learning from Pattern Completion: Self-supervised Controllable Generation}

\author[1,2*\ddag \phantom{\thanks{Equal Contribution.}}]{Zhiqiang Chen}
\author[3{*\dag}\phantom{\thanks{Work done during Internship at Beijing Academy of Artificial Intelligence.}}]{Guofan Fan}
\author[2,1,4{*}]{Jinying Gao}
\author[5]{Lei Ma}
\author[1]{Bo Lei}
\author[1,5]{\\Tiejun Huang}
\author[2,4{\ddag}\phantom{\thanks{Corresponse author. \Letter{ chenzhiqiang@mails.ucas.ac.cn,shan.yu@nlpr.ia.ac.cn.}}}]{Shan Yu}

\affil[1]{Beijing Academy of Artificial Intelligence}
\affil[2]{Institute of Automation, Chinese Academy of Science}
\affil[3]{Xi'an Jiaotong University}
\affil[4]{University of Chinese Academy of Science}
\affil[5]{Peking University}

\begin{document}

\maketitle

\begin{abstract}

 The human brain exhibits a strong ability to spontaneously associate different visual attributes of the same or similar visual scene, such as associating sketches and graffiti with real-world visual objects, usually without supervising information. In contrast, in the field of artificial intelligence, controllable generation methods like ControlNet heavily rely on annotated training datasets such as depth maps, semantic segmentation maps, and poses, which limits the method’s scalability. Inspired by the neural mechanisms that may contribute to the brain’s associative power, specifically the cortical modularization and hippocampal pattern completion, here we propose a self-supervised controllable generation (SCG) framework. Firstly, we introduce an equivariance constraint to promote inter-module independence and intra-module correlation in a modular autoencoder network, thereby achieving functional specialization. Subsequently, based on these specialized modules, we employ a self-supervised pattern completion approach for controllable generation training. Experimental results demonstrate that the proposed modular autoencoder effectively achieves functional specialization, including the modular processing of color, brightness, and edge detection, and exhibits brain-like features including orientation selectivity, color antagonism, and center-surround receptive fields. Through self-supervised training, associative generation capabilities spontaneously emerge in SCG, demonstrating excellent zero-shot generalization ability to various tasks such as superresolution, dehaze and associative or conditional generation on painting, sketches, and ancient graffiti. Compared to the previous representative method ControlNet, our proposed approach not only demonstrates superior robustness in more challenging high-noise scenarios but also possesses more promising scalability potential due to its self-supervised manner. Codes are released on \href{https://github.com/BAAI-Brain-Inspired-Group/OPEN-Vis-ControlSD}{\textbf{\textcolor{gray}{Github}}} and \href{https://gitee.com/chenzq/control-net-main}{\textbf{\textcolor{gray}{Gitee}}}.
  
\end{abstract}

\section{Introduction}
\begin{figure}
  \centering
  \includegraphics[width=0.95\textwidth]{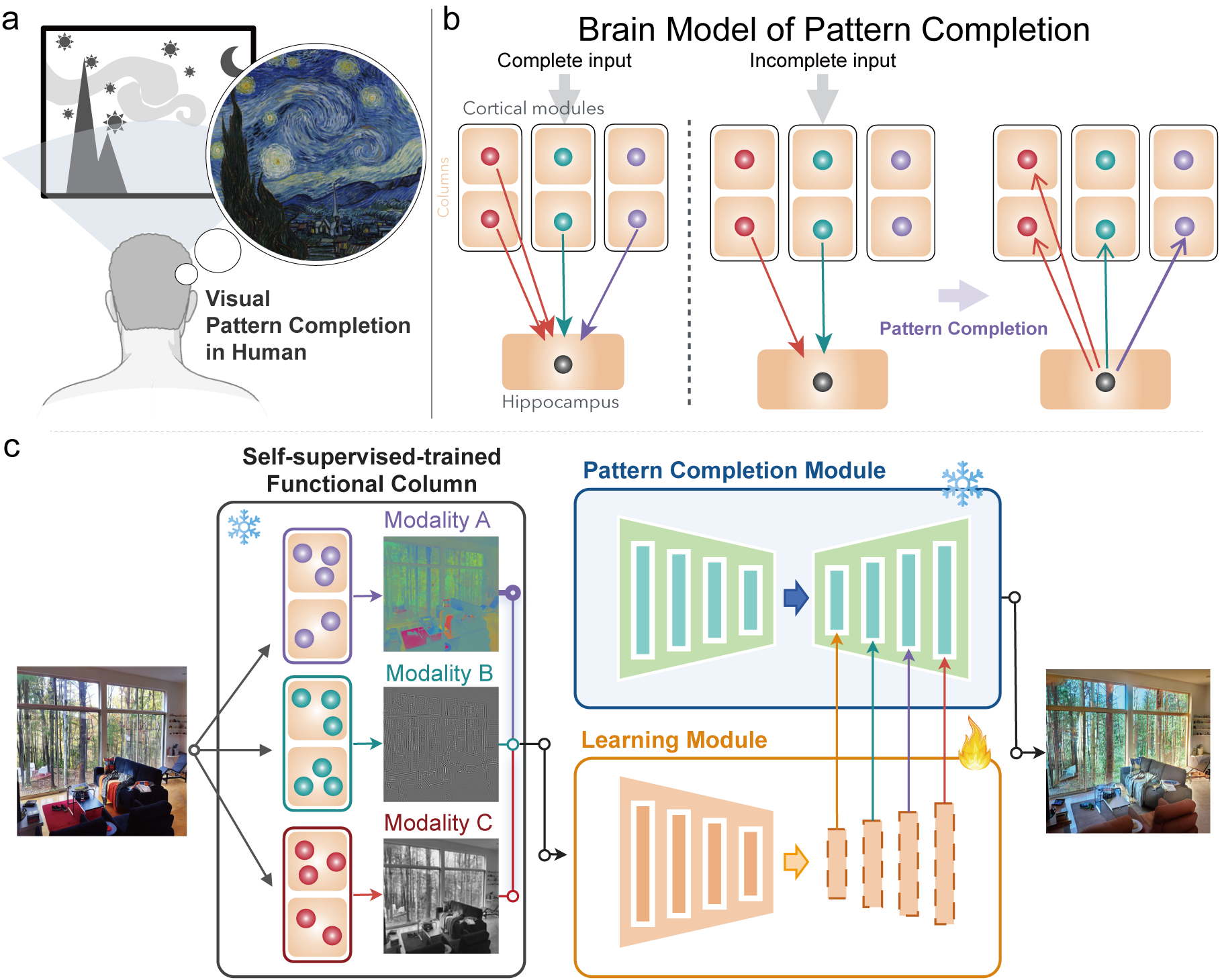}
  \caption{Framework of SCG. SCG has two components. One is to promote the network to spontaneously specialize different functional modules through our designed modular equivariance constraint; The other is to perform self-supervised controllable generation through pattern completion.}\label{fig1_framework}
\end{figure}

In contrast to artificial intelligence, the human brain exhibits a remarkable characteristic: the spontaneous emergence of associative generation\cite{grollier2020neuromorphic,horner2015evidence,guzman2016synaptic}, such as associating real-world visual scenes through pictures, sketches, graffiti, and so on (Figure \ref{fig1_framework}a). With the development of controllable generation technology\cite{Zheng_2023_CVPR,avrahami2023spatext,xu2023magicanimate}, many excellent works have recently been able to achieve similar associative generation or conditional generation capabilities, such as ControlNet\cite{Zhang_2023_ICCV}. However, existing controllable generation models often require supervised training to achieve this function, such as edge maps\cite{Xie_2015_ICCV}, semantic segmentation maps\cite{long2015fully}, depth maps\cite{liu2015deep}, and pose\cite{toshev2014deeppose}. 

The brain’s remarkable ability to generate associations emerges spontaneously and likely stems from two key mechanisms. One is the brain’s modularity in terms of structure and function\cite{gallen2019brain}, enabling it to decouple and separate external stimuli into distinct patterns. This modularity is essential for a range of subsequent cognitive functions, including associative generation. Although previous works such as group convolution\cite{zhang2018shufflenet,ma2018shufflenet}, capsule networks\cite{sabour2017dynamic,hinton2018matrix}, and mixture of experts models\cite{zhou2022mixture,riquelme2021scaling} have designed modular grouping structures, and often show good advantages in model parameter efficiency and computational efficiency, they cannot spontaneously specialize reliable functional modules. Another is the pattern completion ability of hippocampus\cite{sur1990cross}(Figure \ref{fig1_framework}b), such as associating the visual signal of a pepper with its taste, associating a sketch with a real visual scene, etc. If we regard edge maps, semantic segmentation maps, depth maps, etc. as general modalities or patterns, then existing controllable generation works in artificial intelligence can also be regarded as a kind of pattern completion process\cite{chen2023rethinking}. Due to the lack of the brain’s ability to spontaneously and reliably specialize functional modules\cite{chernjavsky1990spontaneous}, controllable generation often adopts a supervised approach. It brings two problems: One is the need for manually defined supervised conditions; The other is the need for a large amount of annotated data. This supervised approach may limit the scalability of associative generation, which is one of the most important capabilities in the era of large models\cite{devlin2018bert,radford2018improving,he2022masked}. Therefore, it is meaningful to enable the network to spontaneously specialize functional modules and perform pattern completion for controllable generation in a self-supervised manner\cite{chrysos2021unsupervised}.


To mimick functional modularity\cite{meunier2010modular,bukach2006beyond,bonhoeffer1991iso} of brain, we incorporate an equivariance constraint\cite{bressloff2002visual,cohen2016group} into the training process of an unsupervised modular autoencoder to enhance the correlation within a module and the independence between the modules, thus promoting the functional specialization. Experiments on multiple datasets including MNIST and ImageNet demonstrate that the proposed equivariance constraint can enable each module to form a closed and complete independent submanifold, resulting in significant and reliable functional specialization similar to biological systems, such as orientation, brightness, and color. As shown in Figure \ref{fig1_framework}c, with the spontaneously specialized functional modules as conditions, we train a self-supervised controllable generation (SCG) adapter based on a pretrained diffusion model to achieve pattern completion and generate images. Experimental results on the COCO dataset show that SCG achieves better structural similarity and semantic similarity compared to ControlNet. At the same time, SCG also has excellent zero-shot generalization in associative generation tasks ranging from oil paintings, ink paintings, sketches, and more challenging ancient graffiti. The generated images both have similar contents and structures to the original images and rich and vivid generated details. Especially in the subjective evaluation of associative generation of oil paintings and ancient graffiti, compared to previous excellent work ControlNet, SCG has a higher or similar winning rate in fidelity and a significantly higher winning rate in aesthetics.

\section{Related Works}
\textbf{Controllable Generation: }Controllable generative models enhance the generation process by incorporating control conditions\cite{Zheng_2023_CVPR,avrahami2023spatext,xu2023magicanimate}, leading to images that more closely align with desired outcomes. Compared to using text as control conditions\cite{li2019controllable,huang2023t2i,kumari2023multi,kang2023scaling,takagi2023high,qu2023layoutllm}, image-based control conditions\cite{chen2018sketchygan,mou2024t2i,Zhang_2023_ICCV,zhao2024uni,ma2024follow} afford finer-grained controllability, which often need annotated paired images for supervised training. While some methods leverage structural similarity for additional constraint\cite{bashkirova2023masksketch,zhu2017unpaired}, enabling style transfer\cite{richardson2021encoding} while maintaining structural consistency, these approaches typically rely on GANs\cite{goodfellow2020generative} rather than stronger diffusion models\cite{ho2020denoising}.

\textbf{Modular Neural Networks: }The most direct approach to modular network design is structural modularity, achieved by techniques such as dividing convolutional kernels into distinct groups within each layer\cite{zhang2018shufflenet,ma2018shufflenet,zhang2017interleaved,sun2018igcv3} or designing specialized blocks\cite{szegedy2015going,szegedy2016rethinking}. Capsule networks\cite{sabour2017dynamic,hinton2018matrix,rajasegaran2019deepcaps,kosiorek2019stacked} and Mixture of Experts (MoE)\cite{zhou2022mixture,riquelme2021scaling} both adopt dynamic routing mechanism between diverse blocks. These methods often improve parameter and computational efficiency but struggle to achieve reliable and stable functional specialization. Group equivariance\cite{cohen2016group,shen2020pdo,weiler2018learning,weiler2019general,hsu1990holographic,chen2024continuous,chen2022sharing} offers a novel approach for functional specialization. By imposing equivariance constraints, the network automatically learns orientation-selective features akin to simple cells in the visual cortex, resulting in clear functional specialization\cite{gao2022learning}. However, this approach is limited to translational equivariance.


\section{Self-supervised Controllable Generation}
The whole framework of our Self-supervised Controllable Generation (SCG) is shown in Figure \ref{fig1_framework}c. This framework comprises two components: a modular autoencoder and a conditional generator for pattern completion. We first train a special autoencoder to encode the input image into disentangled feature spaces or modalities called Modular Autoencoder. Then we use part of the modules as control conditions to complete the missing information and reconstruct the input image. This self-supervised approach does not require manually designed feature extractors or manual annotations, making it easier to train and more scalable compared to supervised approaches. Notable, given the extensive research on conditional generation, we leverage the existing, mature ControlNet\cite{Zhang_2023_ICCV} for pattern completion part. Our core contribution lies in designing a modular autoencoder based on proposed equivariance constraint, successfully enabling the network to spontaneously develop relatively independent and highly complementary modular features. These features are crucial for subsequent conditional generation.



\textbf{Modular Autoencoder: }The main principle of the modular autoencoder is to enhance the independence between modules and the correlation within modules. Inspired by the primary visual cortex of biological systems, we design a modular equivariance constraint to promote functional specialization. Specifically, for a typical autoencoder, it can be formally represented as:
\begin{equation}
    E(I) = f,D(f)=I
\end{equation}
, where $I$ is the input image, $f$ is the latent representation, $E$ and $D$ are the encoder and decoder, respectively. For a modular autoencoder, we define the independence between modules as: 
\begin{equation}
    f=[f^{(0)},f^{(1)},...,f^{(k-1)}], f^{(i)}(L_{\delta}(I))=P_{\delta}(f^{(i)}(I)).
\end{equation}
$f$ can be divided into several modulars as $[f^{(0)},f^{(1)},...,f^{(k-1)}]$, where $f^{(i)}$ is the $i$th modular. $L_{\delta}$ is a general transformation determined by a low dimension parameter $\delta$. $P_{\delta}$ is a prediction function in the latent space with parameter $\delta$. It means that each module can independently predict its state by the specific change $\delta$, which indicates that each module is equivariant on the corresponding change group. For the correlation within a module, we define
\begin{equation}
    f^{(i)}_{n}=P^{(i)}_{\delta^{*}_{mn}}(f^{(i)}_{m}),
\end{equation}
where $f^{(i)}_{n}$ and $f^{(i)}_{m}$ denote one-hot representations with 1 in the $m$th and $n$th dimensions, respectively. For any two dimensions $f^{(i)}_{m}$ and $f^{(i)}_{n}$ in $f^{(i)}$, there always exist corresponding prediction parameters $\delta^{*}_{mn}$ such that above equation holds.

\begin{figure}
\centering
    \includegraphics[width=0.9\linewidth]{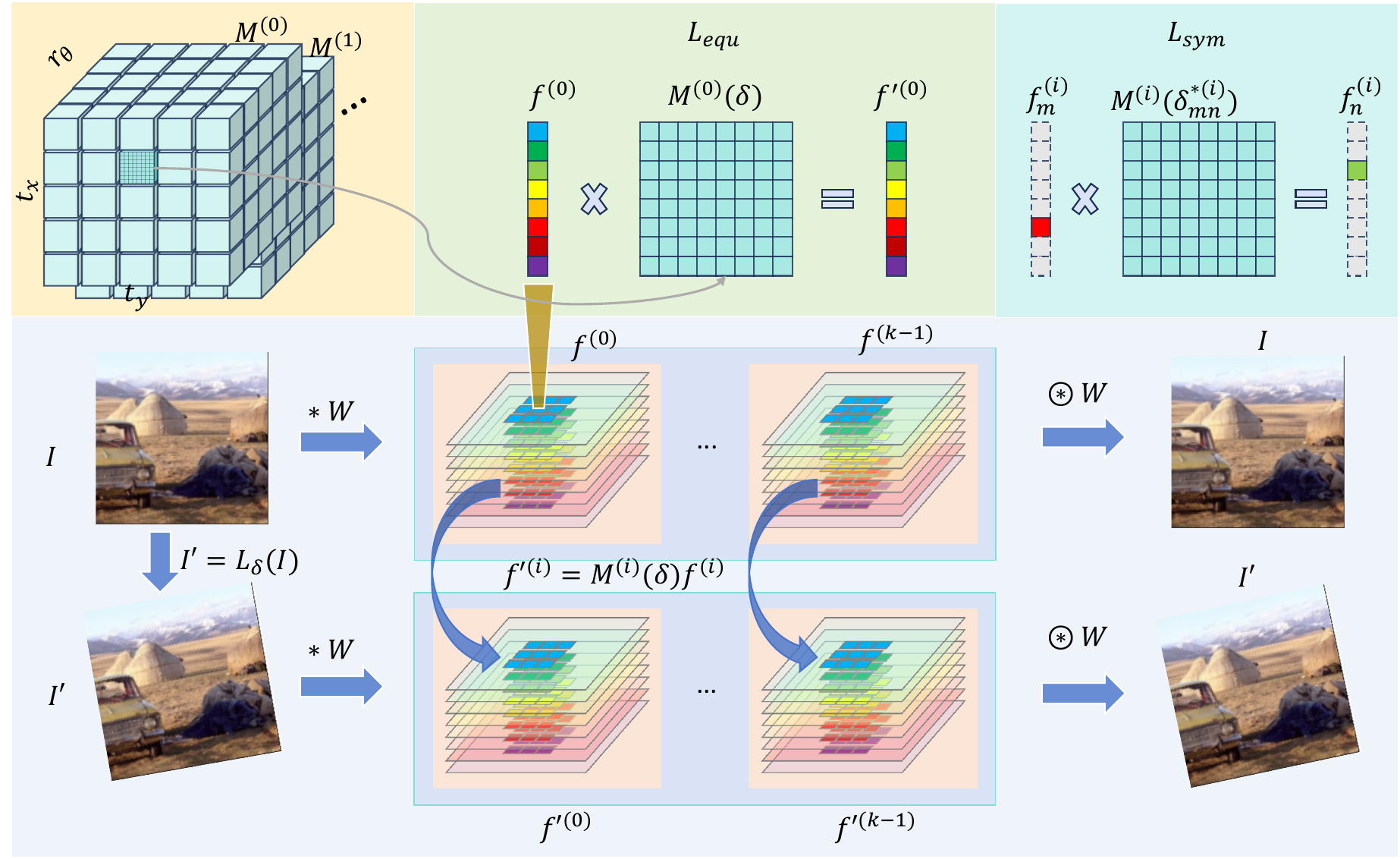}
    \caption{Detail architecture of proposed Modular Autoencoder. The latent space is divided into several modules. We use a prediction matrix $M^{(i)}(\delta)$ to build relationship on latent space between input image pairs. $M^{(i)}$ is the learnable codebooks for each modules.}\label{architecture}
\end{figure}

\textbf{Equivariance constraint: }In practical implementation, we use simple translation and rotation transformations as the change $L_{\delta}$, and a learnable linear transformation matrix $M^{(i)}(\delta)$ as the prediction function. 

As shown in Fig. \ref{architecture}, we utilize convolution operation as the encoder $E$ with kernel $W$. We train the autoencoder on image pairs $I$ and $I'$, where $I'=L_{\delta}(I)$. The image encoding process is denoted as $f=I*W$ and $f'=I'*W$, where $W$ is learnable convolutional kernels, $*$ denotes convolution, and $f$ and $f'$ represent the extracted feature maps. The decoding process is $f \circledast W$ and $f' \circledast W$, where $W$ shares kernels used in the encoding process, and $\circledast$ denotes deconvolution. The objective of the autoencoder is to make the reconstructed image as similar as possible to the original image. So we design the reconstruction loss as follows:
\begin{equation}\label{L1}
  L_{recon} = ||f\circledast W-I||^2 + ||f'\circledast W -I'||^2
\end{equation}

Then, we construct an equivariant loss to drive modules in the networks to be more independent. We divide the convolutional kernels $W$ into $k$ groups: $W=[W^{(0)},W^{(1)},...,W^{(k-1)}]$, and the corresponding feature maps $f$ are also divided into $k$ groups: $f=[f^{(0)},f^{(1)},...,f^{(k-1)}]$. Within each module, we apply an equivariance constraint such that
\begin{equation}\label{L2}
L_{equ}=\sum_{i}||f'^{(i)}-M^{(i)}(\delta)f^{(i)}||^2,i\in\{0,1,...,k-1\}
\end{equation}
, where $M^{(i)}(\delta)$ is a learnable prediction matrix determined by $\delta$. Specifically, $M^{(i)}$ is a 3D matrix codebook indexed by $\delta$ (linear interpolation), where $\delta$ is composed of two translation parameters $t_x$ and $t_y$ and one rotation parameter $r_\theta$. Next, we construct a symmetry loss to enhance the relationship within each module:
\begin{equation}\label{L3_1}
\begin{aligned}
\delta^{(i)*}_{mn}&=\sum_{\delta}{\frac{e^{-M^{(i)}_{mn}(\delta)/T}}{\sum_{\delta'}{e^{-M^{(i)}_{mn}(\delta')/T}}}*\delta}, m,n\in\{0,1,...,l-1\},\\\delta,\delta'&=(t_x,t_y,r_{\theta}),t_x,t_y\in[-0.5s,0.5s),r_{\theta}\in[0,2\pi))\\
L_{sym}&=\sum_{i}\sum_m\sum_n||f^{(i)}_{m}-M^{(i)}(\delta^{(i)*}_{mn})f^{(i)}_{n}||^2,i\in\{0,1,...,k-1\}
\end{aligned}
\end{equation}
where $l$ is the module length, and $M^{(i)}_{mn}(\delta)$ is the $m$ row and $n$ column of the prediction matrix, which indicates the relationship between $m$th and $n$th dimensions module $f^{(i)}$ corresponding to $\delta$. $s$ is the convolution stride. $\delta^{(i)*}_{mn}$ is the best transformation parameter from $n$th dimension to $m$th dimension calculated by a softmax function with temperature $T$. $f^{(i)}_{m}$ and $f^{(i)}_{n}$ are two one-hot representation with 1 in the $m$th and $n$th dimensions, respectively. The goal of the symmetry loss function is to make any two dimensions of the same module symmetric with respect to a certain transformation $\delta^{(i)*}_{mn}$. 

Finally, we take above three losses together as our equivariance constraint:
\begin{equation}\label{L3_3}
L_{EC}=L_{recon} + \lambda_{1}L_{equ} + \lambda_{2}L_{sym},
\end{equation}
where $\lambda_{1}$ and $\lambda_{2}$ are the weights of the equivariant loss and symmetry loss, respectively.

\begin{figure}
  \centering
  \includegraphics[width=0.98\textwidth]{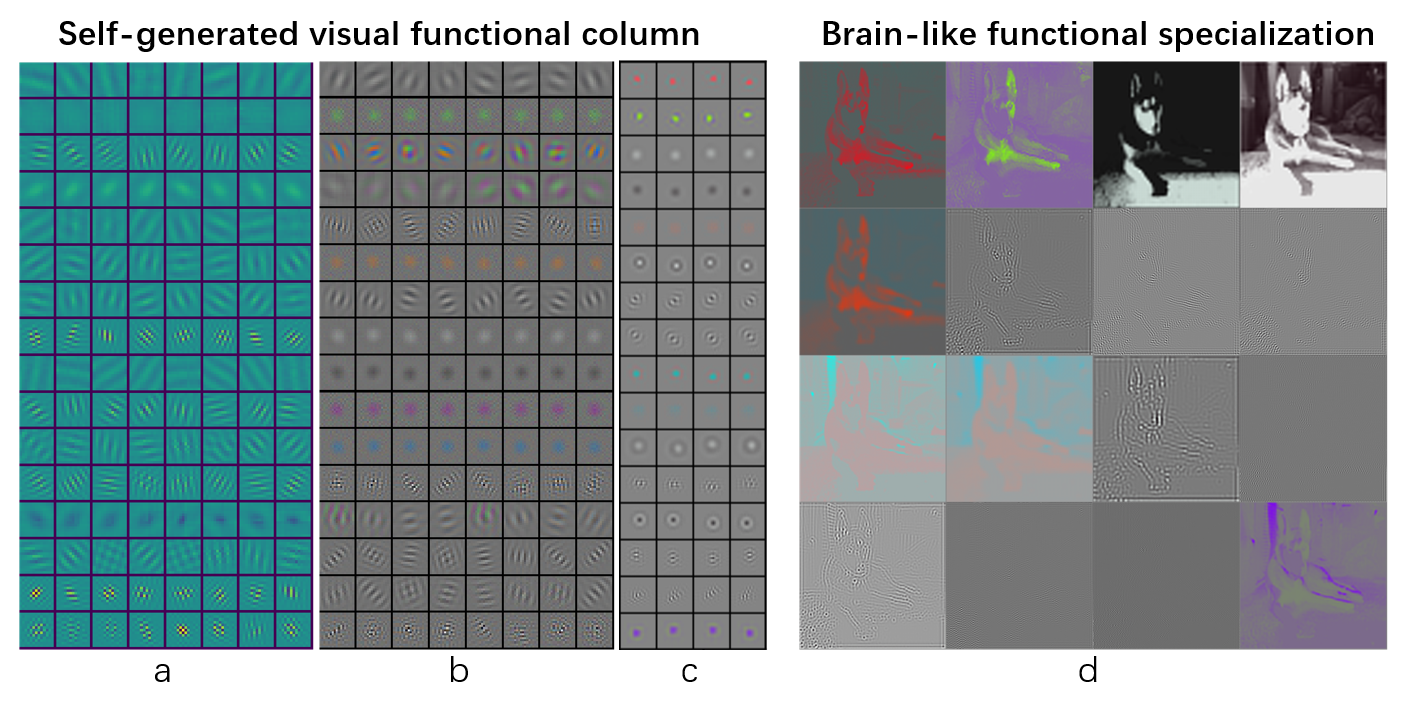}
  \caption{Feature Visualization of modular autoencoder. Each panel shows all features learned by an individual model with multiple modules (one module each row). We trained modular autoencoder with a translation-rotation equivariance constraint on a)MNIST and b)ImageNet, respectively. c) On ImageNet, we also train an autoencoder with an additional translation equivariance constraint besides the translation-rotation equivariance constraint on each module. d) We visualize reconstructed images by features of each module in c.}\label{fig2_ModularDiff}
\end{figure}

\textbf{Pattern Completion: }Based on the modular autoencoder trained in the previous section, we can treat each module as a general modality or pattern and then perform conditional generation through pattern completion. This process can be formulated as:
\begin{equation}
    f=C(f^{(i)},i),
\end{equation}
where $f^{(i)}$ is $i$th module representation of $I$. The pattern completion is learning a completion function $C$ to complete the whole representation $f$. If we have a process of module specialization, then we can train completion function in a self-supervised manner, similar to how humans do.

In practice, we implement the pattern completion process based on the modular autoencoder designed in the previous section, referring to the existing outstanding controllable generation work ControlNet. Specifically, if we use module $f^{i}$ as a control condition, then the overall learning objective of self-supervised controllable generation based on ControlNet can be expressed as:
 \begin{equation}
     L_{MC}=\mathbb{E}_{z_0,t,c_t,f^{(i)},\epsilon \sim N(0,1)}[||\epsilon - \epsilon_{\theta}(z_t,t,c_t,f^{(i)})||^2].
 \end{equation}
 $z_0$ is the latent representation for the input image and $z_t$ is the noisy image at time step $t$. $c_t$ is the text prompt, and $f^i$ is a module of the pre-trained modular autoencoder. Image diffusion algorithms learn a network $\epsilon_{\theta}$ to predict the noise $\epsilon$. The generation process conditioned by parts of the module's information can be regarded as a pattern completion process. Thus, we achieve self-supervised controllable generation through an automated modularization and pattern completion process.

\section{Experiments and Results}\label{experiment}



\textbf{Modular Autoencoder: }We trained our modular autoencoder mainly on two typical datasets. One is MNIST, which is a small dataset with gray digits. The other is ImageNet, which is a relatively large dataset with color natural images. On both MNIST and ImageNet, modular autoencoder can achieve obvious function specialization as shown in Figure \ref{fig2_ModularDiff}. Many learned kernels have significantly gabor-like orientation selectivities as shown in Figure \ref{fig2_ModularDiff}a and b, which is a brain-like characteristic. On datasets consisting of gray images, such as MNIST, modules primarily specialize into selectivity for different spatial frequency orientations, with each module having the same spatial frequency and complete orientation selectivity (Figure \ref{fig2_ModularDiff}a). On datasets consisting of natural color images, modules not only specialize into selectivity for different spatial frequency orientations but also specialize into modules that are sensitive to color and brightness (Figure \ref{fig2_ModularDiff}b). Unlike imposing translation equivariance constraints on each kernel, Figure \ref{fig2_ModularDiff}c imposes both translation equivariant and trans-rot equivariance constraints on the entire module. This allows it to learn brain-like center-surround receptive fields, and generate center-surround antagonistic and color-antagonistic modules(See in \ref{appendix_color_centersurround}), which are considered to be more robust to noise. Similarly, Figure \ref{fig2_ModularDiff}c also forms obvious functional specialization such as color, brightness, and edges, which can also be observed from the reconstruction visualization of each module in Figure \ref{fig2_ModularDiff}d. See more experiment, training setup, dataset and ablation study details in Appendix A.4.1, A.4.2, A.4.3 and A.2.

\begin{figure}
  \centering
  \includegraphics[width=0.96\textwidth]{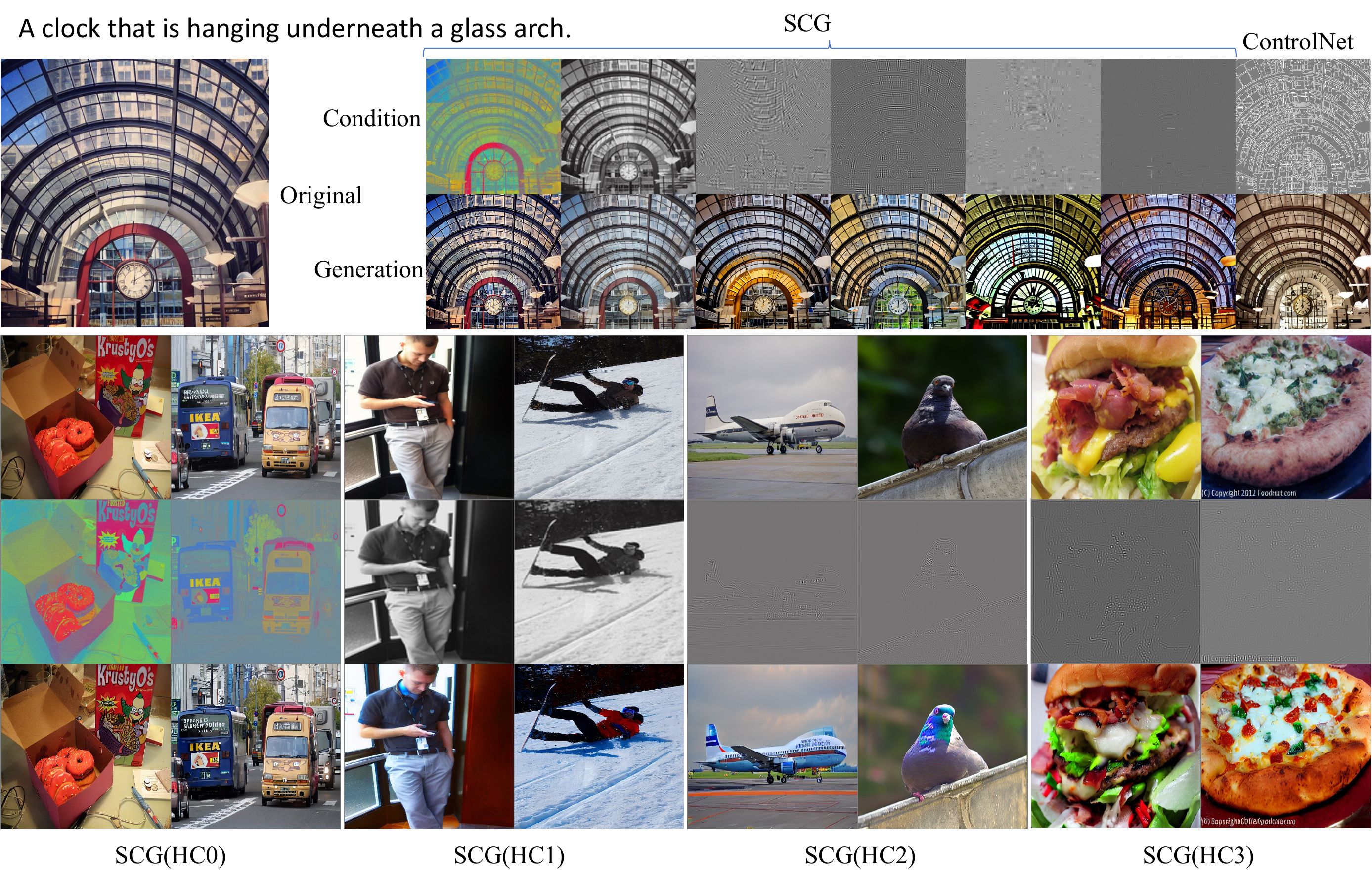}
  \caption{Images generated by SCG in MS-COCO. The upper part shows an image randomly selected in MS-COCO with a text prompt. On the right show the condition images extracted from our modular autocoder and the corresponding generated images. The last column is a generated image by ControlNet conditioned by the canny edge. The bottom part shows more generated images. The three row images are original, condition and generated images, respectively.(See more in Figure \ref{appendix_cocotr} and \ref{appendix_cocoval})}\label{fig3_cocoval}
\end{figure}
\begin{figure}
\begin{minipage}{0.68\linewidth}
  \renewcommand{\arraystretch}{1.0}
  \setlength\tabcolsep{2.5pt}
  \centering
  \captionof{table}{Qualitative evaluation on MS-COCO. $\uparrow$ means that higher is better, and $\downarrow$ means the opposite. g and c means gray images and color images, respectively.}
  \label{table1_comparison}
  \begin{tabular}{l|ccccccc}
    \toprule
                            & \small{}   & \multicolumn{6}{c}{SCG}\\
                            \cmidrule(r){3-8}
    Metrics                 & \small{ControlNet}    & HC0                     &HC1     & HC2    & HC3     & HC4     & HC5 \\
    \midrule
    PSNR(g)$\uparrow$    & 10.7      & 19.2           & \textbf{19.9}   & 17.9   & 14.8    & 11.5    & 11.2 \\
    PSNR(c)$\uparrow$   & 10.3      & \textbf{19.0}           & 16.8   & 15.6   & 13.7    & 10.8    & 10.7 \\
    SSIM(g)$\uparrow$    & 0.125     & 0.486          & \textbf{0.519}  & 0.445  & 0.352   & 0.191   & 0.177 \\
    SSIM(c)$\uparrow$   & 0.128     & \textbf{0.490}          & 0.443  & 0.388  & 0.320   & 0.179   & 0.170 \\
    FID\cite{heusel2017gans}$\downarrow$         & 13.7      & \textbf{8.5}            & 11.0   & 11.2   & 12.1    & 17.1    & 14.9  \\
    CLIP\cite{radford2021learning}$\uparrow$          & 0.846     & \textbf{0.958}          & 0.898  & 0.890  & 0.878   & 0.812   & 0.830 \\
    \bottomrule
  \end{tabular}
\end{minipage}
\begin{minipage}{0.3\linewidth}
  \centering
  \includegraphics[width=\textwidth]{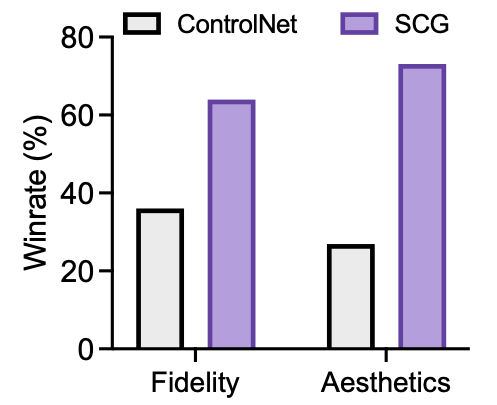}
  \caption{Subjective evaluation on zero-shot oil painting accociation generation.}\label{oilpaintingeval}
\end{minipage}
\end{figure}

\textbf{Self-supervised Controllable Generation on MS-COCO\cite{lin2014microsoft}: }We implement our SCG based on our trained modular autoencoder in Figure \ref{fig2_ModularDiff}c (see more details in \ref{appendix_training_setup}). Specifically, 
we divided the 16 modules into 6 groups named from HC0 to HC5 (see details in \ref{appendix_experiment_details}), and generated images conditioned by them respectively as shown in Figure \ref{fig3_cocoval}. The original images with text prompts are randomly selected on MS-COCO in the first line. The condition images are in the second line. In Figure \ref{fig3_cocoval}, some modules provide the color information, some provide the brightness information, and some provide the edge information with different spatial frequencies. The last condition image is the edge map extracted by the Canny detector, which is used in ControlNet. The third line is the generated images by different conditions. With incomplete control information as input, the diffusion model learns to complete the missing information to obtain a complete image. Due to the different control information, the generated images also have their own characteristics. The bottom shows more generation results. The three lines are original images, condition images, and generated images, respectively. HC0 mainly extracts color information, and the generated image is also closely similar to the original image. HC1 provides brightness information but lacks color information. The generated image is recolored with a similar brightness structure to the original image. HC2 and HC3 provide structural information such as edge and corner but lack color, brightness and detail information. The corresponding generated images can recolor images and regenerate many vivid and reasonable details. A quantitative analysis is performed on the validation set of MS-COCO in Table \ref{table1_comparison}. We use two traditional similarity metrics PSNR and SSIM and two semantic-oriented metrics FID and CLIP to measure the similarity between the generated image and the original image. For PSNR and SSIM, we measure it in gray and color images, respectively. As in Table \ref{table1_comparison}, in terms of the traditional metrics PSNR and SSIM, the proposed SCGs conditioned by the autonomously differentiated modules all achieve higher similarities than ControlNet conditioned by artificial Canny edge and show a decreasing trend from HC0 to HC5. In the gray images that does not consider color information, the image generated by HC1 that provides brightness information is higher than that generated by HC0 that provides color information in terms of PSNR and SSIM similarity. In terms of semantic similarity, HC0 is the best in FID and CLIP similarity, and shows a gradual decrease in similarity from HC0 to HC3, and all of them are better than the similarity of ControlNet. It also demonstrates that the information provided by HC0 to HC3 can better reflect the semantic structure than HC4 and HC5, so the generated images are also more semantically similar to the original images.

\section{Conditional Associative Generation}

Associating real-world scenes from different styles of paintings and abstract graffiti is a capability that everyone possesses. And this ability does not require supervised training, it is completely self-emergent and has zero-shot generalization ability. The proposed SCG can also emerge with conditional and association generation ability. We use SCG trained on COCO dataset to test its zero-shot generalization capabilities on conditional and associative generation with sketches, oil paintings, wash and ink paintings and more challenging ancient graffiti on rock.

\begin{figure}
  \centering
  \includegraphics[width=0.98\textwidth]{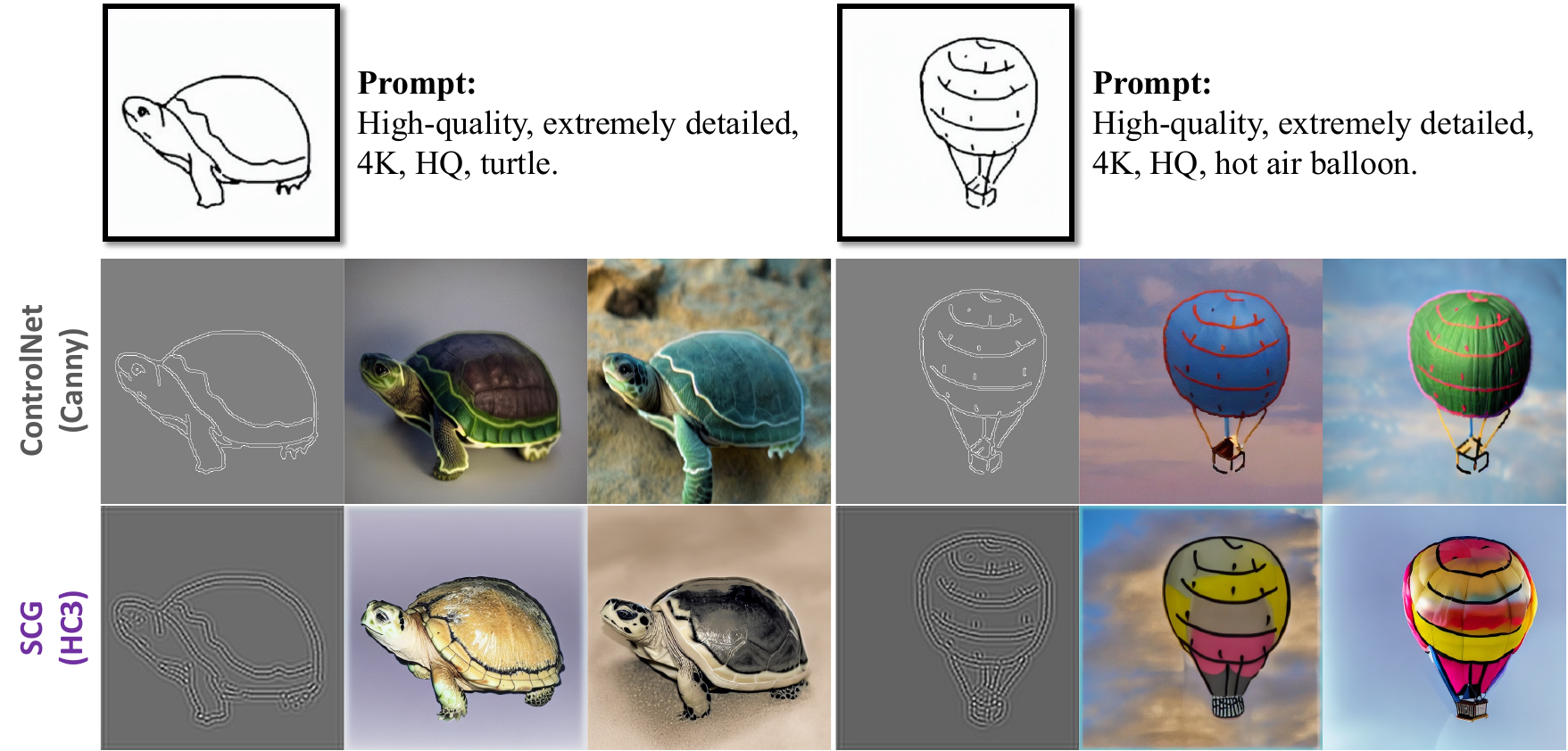}
  \caption{Association generation on manual sketches. The original sketches are from ControlNet\cite{Zhang_2023_ICCV}.}\label{fig7_scribble}
\end{figure}

\textbf{Sketches: }We first test it on manual sketches. We tested both ControlNet conditioned by Canny edges and SCG conditioned by HC3. In Figure \ref{fig7_scribble}, ControlNet can generate images from sketches in excellent fidelity and aesthetics to the original sketches. Basing on the Canny edge detector, ControlNet possesses a natural advantage when dealing with sketches, resulting in high-quality generation outcomes. For SCG, sketch is an entirely novel domain with a significant divergence from the training dataset distribution. Nevertheless, SCG can still generate images with remarkable fidelity and aesthetics by an automatic specialized feature extractor, demonstrating its excellent generalization capabilities.




\begin{figure}
  \centering
  \includegraphics[width=0.98\textwidth]{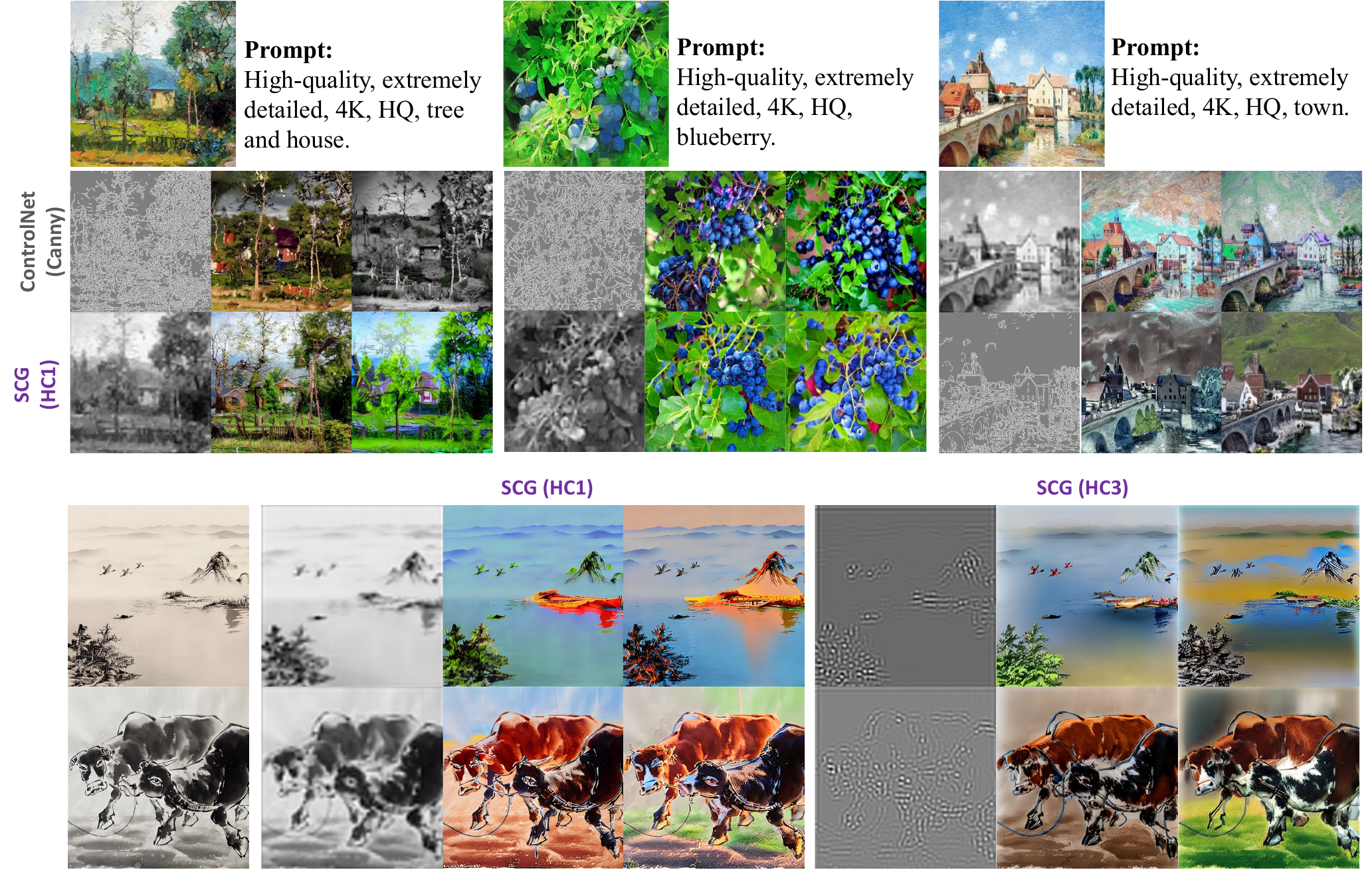}
  \caption{Associative generation on oil painting (top) and wash and ink painting (bottom). (See more generation results in Figure \ref{appdix_fig1_oil} and Figure \ref{appdix_fig1_shuimo})}\label{fig5_painting}
\end{figure}

\textbf{Painting: }We also tested two typical painting styles: Western-style oil paintings and Eastern-style wash and ink paintings. Figure \ref{fig5_painting} shows the results of associative generation on oil paintings on the top part. The first row is the original oil painting images with text prompts, and the second row is the condition image Canny edge and generated images of ControlNet. Compared to clean sketches, oil paintings inevitably contain content-irrelevant textures that can interfere with the feature extraction process of the Canny edge detector. Therefore, the generated images will also be affected to some extent, resulting some detailed textures not that natural and beautiful. For our SCG in the third row, we use HC1 as condition, which is sensitive to brightness. Our brain-like HC1 has a better noise suppression effect, and the generated image details are more vivid and beautiful. For quantitative evaluation in Figure \ref{oilpaintingeval}, we conduct subjective evaluations of the generated images in terms of fidelity and aesthetics (see details in Appendix A.4.4). As shown in Figure \ref{oilpaintingeval}, compared to ControlNet, SCG has significantly higher winning rates in both fidelity and aesthetics. In addition to Western-style oil paintings, we also test Eastern-style wash and ink paintings. In the bottom part of Figure \ref{fig5_painting}, the first column is the original wash and ink paintings, and we generated 2 images each with HC1 and HC3 as conditions, respectively. The generated images have similar content and structure to the original paintings. At the same time, SCG recolors the painting and generates natural and reasonable textures.



\begin{figure}
  \centering
  \includegraphics[width=0.98\textwidth]{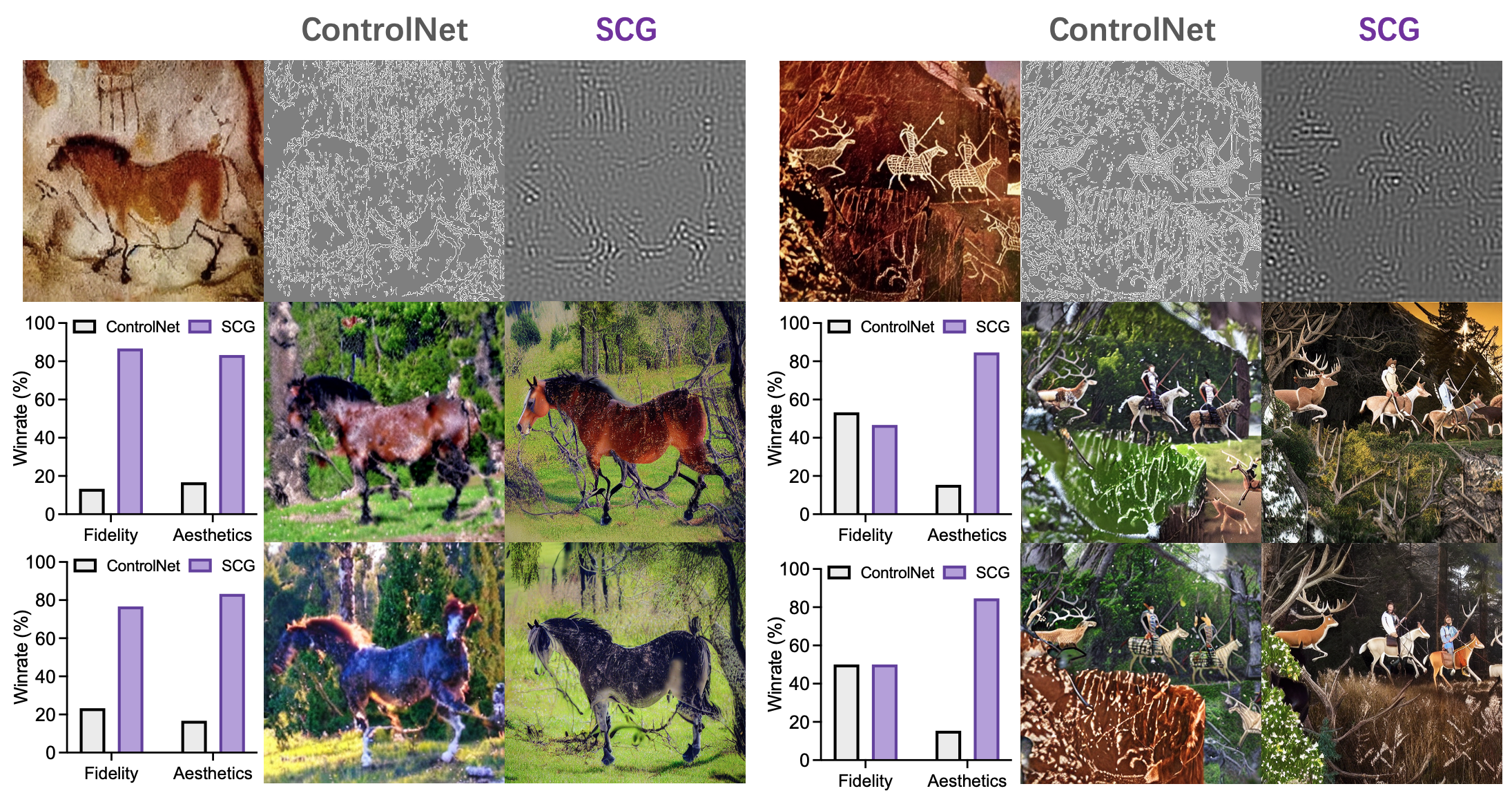}
  \caption{Association generation on ancient graffiti on rock. (See more generations in Figure \ref{appendix_bihua})}\label{fig5_bihua}
\end{figure}
\textbf{Ancient Graffiti: }In contrast to clean sketch without noise and painting, the associative generation of ancient graffiti on rock is a more challenging task due to the inherent high noise levels resulting from their presence on natural rock surfaces over thousands of years. Similarly, we also test the results of ControlNet based on Canny and SCG based on HC3 in Figure \ref{fig5_bihua}, respectively. Due to high-level noise, the artificially designed Canny edge detector is seriously disturbed by noise, resulting in relatively cluttered edge maps. Compared to Canny, our learned modules exhibit strong robustness to noise. For quantitative comparison, we conduct a subjective evaluation on fidelity and aesthetics. For the first graffiti of a horse, the images generated by the proposed SCG significantly outperformed those from the Canny-based ControlNet in terms of both fidelity and aesthetics. For fidelity, the four vertical lines above the horse’s back are not reflected in the image generated by ControlNet, but four trees corresponding to the lines are generated by SCG. The overall clarity of SCG is also better than ControlNet. In the hunting scene on the right, Canny edge captures many noise lines and spots on the rock, which is also reflected in its generated image. Although it makes the generated images more similar to the original image, these noisy pieces of information are content-irrelevant and can significantly reduce the aesthetics of the image. Our feature suppresses the noise effectively while extracting overall structures and contents. Therefore, the generated images are more natural and aesthetically pleasing in terms of detail while being structurally similar to the original graffiti. In subjective evaluation, SCG has a similar winning rate on fidelity and a significantly higher winning rate on aesthetics. These demonstrate that the human-like self-supervised pattern completion approach can effectively learn controllable generation capabilities, while exhibiting exceptional generalization ability and robustness in more challenging scenarios.

We also add comparisons to other conditions besides Canny, including depth maps, normal directions, and semantic segmentation in Appendix Fig. \ref{figr1}. These methods require supervised pre-trained feature extractors, which are sensitive to data distribution. We also tested more tasks such as super-resolution and dehazing, as shown in Appendix Fig. \ref{figr2}, and found that SCG, in addition to spontaneously emerging conditional generation capabilities for oil paintings, ink paintings, ancient graffiti, sketches, etc., still zero-shot generalized abilities of super-resolution, dehazing, and controllable generation under line art conditions. 


\section{Discussion and Limitation}\label{limitation}

The proposed equivariance constraint enables the network to spontaneously specialize into modules with distinct functionalities. Further experiments reveal features within each module exhibit similar spatial frequency selectivity, while different modules possess distinct spatial frequency preferences(see in \ref{appendix_tuning_curve}). Within each module, features exhibit varying orientation selectivity, covering the entire orientation space through a population coding scheme, forming a continuous and closed independent manifold space(see in \ref{appendix_population_submanifold}). These findings suggest the proposed equivariance constraint effectively promotes both intra-module correlation and inter-module independence, facilitating functional specialization. The brain-inspired mechanisms and emergence of brain-like phenomena provide a novel perspective for gaining deeper insights into the learning mechanism of the brain. However, the current equivariance constraint is a simplified and preliminary version, resulting in only low-level feature specialization. While more features, such as depth (parallax), motion(e.g., optical flow), and semantic-oriented contours and instance segmentation, remain unlearned, they hold significant potential to further enhance the controllability or more various tasks. Nevertheless, it validates the efficacy of using equivariance constraints to achieve modularization. Future endeavors will focus on inducing richer functional specialization encompassing semantic hierarchies.

Inspired by the hippocampal pattern completion in the brain, we propose a self-supervised controllable generative framework based on automatically specialized functional modules. Since our modular autoencoder learns a set of complementary and relatively complete modules, most conditional generation or reconstruction tasks can be considered as scenarios where information in one or more modules is missing or damaged. This is also the source of SCG’s zero-shot generalization capabilities such as sketch-conditioned generation, super-resolution, dehazing, etc.. While fully self-supervised training can exhibit impressive emergent capabilities and demonstrate strong generalization across data distributions and task variations, it still falls short compared to dedicated supervised methods in specific tasks. Leveraging our self-supervised model as a pre-trained model and then fine-tuning it on labeled data for specific downstream tasks can significantly improve performance on those tasks. We leave these to future works.


\section{Conclusion}
 Inspired by the visual hypercolumn, we propose a modular autoencoder framework with equivariance constraints to automatically achieve functional specialization by promoting inter-module independence and intra-module correlation. Experimental results on multiple datasets demonstrate the effectiveness of this approach in generating functional specialization, exhibiting characteristics reminiscent of the biological primary visual cortex.
Building upon these differentiated modules, we develop a self-supervised controllable generative framework inspired by hippocampal modal completion. Experiments reveal that this framework can spontaneously zero-shot emerge human-like associative generation, conditional generation by sketch and line art, superresolution, dehaze. And it exhibits strong generalization abilities even in scenarios involving significant distributional differences, such as associating sketches and ancient graffiti with natural visual scenes. 

\section{Acknowledge}
This work was supported by National Key R\&D Program of China(2022ZD0116313), International Partnership Program of Chinese Academy of Sciences(173211KYSB2020002), Beijing Natural Science Foundation (No. JQ24023) and the Beijing Municipal Science \& Technology Commission Project (No. Z231100006623010).

\bibliography{neurips}

\begin{thebibliography}{10}

\bibitem{avrahami2023spatext}
Omri Avrahami, Thomas Hayes, Oran Gafni, Sonal Gupta, Yaniv Taigman, Devi Parikh, Dani Lischinski, Ohad Fried, and Xi~Yin.
\newblock Spatext: Spatio-textual representation for controllable image generation.
\newblock In {\em Proceedings of the IEEE/CVF Conference on Computer Vision and Pattern Recognition}, pages 18370--18380, 2023.

\bibitem{bashkirova2023masksketch}
Dina Bashkirova, Jos{\'e} Lezama, Kihyuk Sohn, Kate Saenko, and Irfan Essa.
\newblock Masksketch: Unpaired structure-guided masked image generation.
\newblock In {\em Proceedings of the IEEE/CVF Conference on Computer Vision and Pattern Recognition}, pages 1879--1889, 2023.

\bibitem{bonhoeffer1991iso}
Tobias Bonhoeffer and Amiram Grinvald.
\newblock Iso-orientation domains in cat visual cortex are arranged in pinwheel-like patterns.
\newblock {\em Nature}, 353(6343):429--431, 1991.

\bibitem{bressloff2002visual}
Paul~C Bressloff and Jack~D Cowan.
\newblock The visual cortex as a crystal.
\newblock {\em Physica D: Nonlinear Phenomena}, 173(3-4):226--258, 2002.

\bibitem{bukach2006beyond}
Cindy~M Bukach, Isabel Gauthier, and Michael~J Tarr.
\newblock Beyond faces and modularity: the power of an expertise framework.
\newblock {\em Trends in cognitive sciences}, 10(4):159--166, 2006.

\bibitem{chen2023rethinking}
Jiaxuan Chen, Yu~Qi, and Gang Pan.
\newblock Rethinking visual reconstruction: Experience-based content completion guided by visual cues.
\newblock 2023.

\bibitem{chen2018sketchygan}
Wengling Chen and James Hays.
\newblock Sketchygan: Towards diverse and realistic sketch to image synthesis.
\newblock In {\em Proceedings of the IEEE conference on computer vision and pattern recognition}, pages 9416--9425, 2018.

\bibitem{chen2024continuous}
Zhiqiang Chen, Yang Chen, Xiaolong Zou, and Shan Yu.
\newblock Continuous rotation group equivariant network inspired by neural population coding.
\newblock In {\em Proceedings of the AAAI Conference on Artificial Intelligence}, volume~38, pages 11462--11470, 2024.

\bibitem{chen2022sharing}
Zhiqiang Chen, Ting-Bing Xu, Jinpeng Li, and Huiguang He.
\newblock Sharing weights in shallow layers via rotation group equivariant convolutions.
\newblock {\em Machine Intelligence Research}, 19(2):115--126, 2022.

\bibitem{chernjavsky1990spontaneous}
Alex Chernjavsky and John Moody.
\newblock Spontaneous development of modularity in simple cortical models.
\newblock {\em Neural Computation}, 2(3):334--354, 1990.

\bibitem{chrysos2021unsupervised}
Grigorios~G Chrysos, Jean Kossaifi, Zhiding Yu, and Anima Anandkumar.
\newblock Unsupervised controllable generation with self-training.
\newblock In {\em 2021 International Joint Conference on Neural Networks (IJCNN)}, pages 1--8. IEEE, 2021.

\bibitem{cohen2016group}
Taco Cohen and Max Welling.
\newblock Group equivariant convolutional networks.
\newblock In {\em International conference on machine learning}, pages 2990--2999. PMLR, 2016.

\bibitem{devlin2018bert}
Jacob Devlin, Ming-Wei Chang, Kenton Lee, and Kristina Toutanova.
\newblock Bert: Pre-training of deep bidirectional transformers for language understanding.
\newblock {\em arXiv preprint arXiv:1810.04805}, 2018.

\bibitem{gallen2019brain}
Courtney~L Gallen and Mark D’Esposito.
\newblock Brain modularity: a biomarker of intervention-related plasticity.
\newblock {\em Trends in cognitive sciences}, 23(4):293--304, 2019.

\bibitem{gao2022learning}
Ruiqi Gao, Jianwen Xie, Siyuan Huang, Yufan Ren, Song-Chun Zhu, and Ying~Nian Wu.
\newblock Learning v1 simple cells with vector representation of local content and matrix representation of local motion.
\newblock In {\em Proceedings of the AAAI Conference on Artificial Intelligence}, volume~36, pages 6674--6684, 2022.

\bibitem{goodfellow2020generative}
Ian Goodfellow, Jean Pouget-Abadie, Mehdi Mirza, Bing Xu, David Warde-Farley, Sherjil Ozair, Aaron Courville, and Yoshua Bengio.
\newblock Generative adversarial networks.
\newblock {\em Communications of the ACM}, 63(11):139--144, 2020.

\bibitem{grollier2020neuromorphic}
Julie Grollier, Damien Querlioz, KY~Camsari, Karin Everschor-Sitte, Shunsuke Fukami, and Mark~D Stiles.
\newblock Neuromorphic spintronics.
\newblock {\em Nature electronics}, 3(7):360--370, 2020.

\bibitem{guzman2016synaptic}
Segundo~Jose Guzman, Alois Schl{\"o}gl, Michael Frotscher, and Peter Jonas.
\newblock Synaptic mechanisms of pattern completion in the hippocampal ca3 network.
\newblock {\em Science}, 353(6304):1117--1123, 2016.

\bibitem{he2022masked}
Kaiming He, Xinlei Chen, Saining Xie, Yanghao Li, Piotr Doll{\'a}r, and Ross Girshick.
\newblock Masked autoencoders are scalable vision learners.
\newblock In {\em Proceedings of the IEEE/CVF conference on computer vision and pattern recognition}, pages 16000--16009, 2022.

\bibitem{heusel2017gans}
Martin Heusel, Hubert Ramsauer, Thomas Unterthiner, Bernhard Nessler, and Sepp Hochreiter.
\newblock Gans trained by a two time-scale update rule converge to a local nash equilibrium.
\newblock {\em Advances in neural information processing systems}, 30, 2017.

\bibitem{hinton2018matrix}
Geoffrey~E Hinton, Sara Sabour, and Nicholas Frosst.
\newblock Matrix capsules with em routing.
\newblock In {\em International conference on learning representations}, 2018.

\bibitem{ho2020denoising}
Jonathan Ho, Ajay Jain, and Pieter Abbeel.
\newblock Denoising diffusion probabilistic models.
\newblock {\em Advances in neural information processing systems}, 33:6840--6851, 2020.

\bibitem{horner2015evidence}
Aidan~J Horner, James~A Bisby, Daniel Bush, Wen-Jing Lin, and Neil Burgess.
\newblock Evidence for holistic episodic recollection via hippocampal pattern completion.
\newblock {\em Nature communications}, 6(1):7462, 2015.

\bibitem{hsu1990holographic}
K-Y Hsu, H-Y Li, and Demetri Psaltis.
\newblock Holographic implementation of a fully connected neural network.
\newblock {\em Proceedings of the IEEE}, 78(10):1637--1645, 1990.

\bibitem{huang2023t2i}
Kaiyi Huang, Kaiyue Sun, Enze Xie, Zhenguo Li, and Xihui Liu.
\newblock T2i-compbench: A comprehensive benchmark for open-world compositional text-to-image generation.
\newblock {\em Advances in Neural Information Processing Systems}, 36:78723--78747, 2023.

\bibitem{kang2023scaling}
Minguk Kang, Jun-Yan Zhu, Richard Zhang, Jaesik Park, Eli Shechtman, Sylvain Paris, and Taesung Park.
\newblock Scaling up gans for text-to-image synthesis.
\newblock In {\em Proceedings of the IEEE/CVF Conference on Computer Vision and Pattern Recognition}, pages 10124--10134, 2023.

\bibitem{kosiorek2019stacked}
Adam Kosiorek, Sara Sabour, Yee~Whye Teh, and Geoffrey~E Hinton.
\newblock Stacked capsule autoencoders.
\newblock {\em Advances in neural information processing systems}, 32, 2019.

\bibitem{kumari2023multi}
Nupur Kumari, Bingliang Zhang, Richard Zhang, Eli Shechtman, and Jun-Yan Zhu.
\newblock Multi-concept customization of text-to-image diffusion.
\newblock In {\em Proceedings of the IEEE/CVF Conference on Computer Vision and Pattern Recognition}, pages 1931--1941, 2023.

\bibitem{li2019controllable}
Bowen Li, Xiaojuan Qi, Thomas Lukasiewicz, and Philip Torr.
\newblock Controllable text-to-image generation.
\newblock {\em Advances in neural information processing systems}, 32, 2019.

\bibitem{lin2014microsoft}
Tsung-Yi Lin, Michael Maire, Serge Belongie, James Hays, Pietro Perona, Deva Ramanan, Piotr Doll{\'a}r, and C~Lawrence Zitnick.
\newblock Microsoft coco: Common objects in context.
\newblock In {\em Computer Vision--ECCV 2014: 13th European Conference, Zurich, Switzerland, September 6-12, 2014, Proceedings, Part V 13}, pages 740--755. Springer, 2014.

\bibitem{liu2015deep}
Fayao Liu, Chunhua Shen, and Guosheng Lin.
\newblock Deep convolutional neural fields for depth estimation from a single image.
\newblock In {\em Proceedings of the IEEE conference on computer vision and pattern recognition}, pages 5162--5170, 2015.

\bibitem{long2015fully}
Jonathan Long, Evan Shelhamer, and Trevor Darrell.
\newblock Fully convolutional networks for semantic segmentation.
\newblock In {\em Proceedings of the IEEE conference on computer vision and pattern recognition}, pages 3431--3440, 2015.

\bibitem{ma2018shufflenet}
Ningning Ma, Xiangyu Zhang, Hai-Tao Zheng, and Jian Sun.
\newblock Shufflenet v2: Practical guidelines for efficient cnn architecture design.
\newblock In {\em Proceedings of the European conference on computer vision (ECCV)}, pages 116--131, 2018.

\bibitem{ma2024follow}
Yue Ma, Yingqing He, Xiaodong Cun, Xintao Wang, Siran Chen, Xiu Li, and Qifeng Chen.
\newblock Follow your pose: Pose-guided text-to-video generation using pose-free videos.
\newblock In {\em Proceedings of the AAAI Conference on Artificial Intelligence}, volume~38, pages 4117--4125, 2024.

\bibitem{meunier2010modular}
David Meunier, Renaud Lambiotte, and Edward~T Bullmore.
\newblock Modular and hierarchically modular organization of brain networks.
\newblock {\em Frontiers in neuroscience}, 4:7572, 2010.

\bibitem{mou2024t2i}
Chong Mou, Xintao Wang, Liangbin Xie, Yanze Wu, Jian Zhang, Zhongang Qi, and Ying Shan.
\newblock T2i-adapter: Learning adapters to dig out more controllable ability for text-to-image diffusion models.
\newblock In {\em Proceedings of the AAAI Conference on Artificial Intelligence}, volume~38, pages 4296--4304, 2024.

\bibitem{qu2023layoutllm}
Leigang Qu, Shengqiong Wu, Hao Fei, Liqiang Nie, and Tat-Seng Chua.
\newblock Layoutllm-t2i: Eliciting layout guidance from llm for text-to-image generation.
\newblock In {\em Proceedings of the 31st ACM International Conference on Multimedia}, pages 643--654, 2023.

\bibitem{radford2021learning}
Alec Radford, Jong~Wook Kim, Chris Hallacy, Aditya Ramesh, Gabriel Goh, Sandhini Agarwal, Girish Sastry, Amanda Askell, Pamela Mishkin, Jack Clark, et~al.
\newblock Learning transferable visual models from natural language supervision.
\newblock In {\em International conference on machine learning}, pages 8748--8763. PMLR, 2021.

\bibitem{radford2018improving}
Alec Radford, Karthik Narasimhan, Tim Salimans, Ilya Sutskever, et~al.
\newblock Improving language understanding by generative pre-training.
\newblock 2018.

\bibitem{rajasegaran2019deepcaps}
Jathushan Rajasegaran, Vinoj Jayasundara, Sandaru Jayasekara, Hirunima Jayasekara, Suranga Seneviratne, and Ranga Rodrigo.
\newblock Deepcaps: Going deeper with capsule networks.
\newblock In {\em Proceedings of the IEEE/CVF conference on computer vision and pattern recognition}, pages 10725--10733, 2019.

\bibitem{richardson2021encoding}
Elad Richardson, Yuval Alaluf, Or~Patashnik, Yotam Nitzan, Yaniv Azar, Stav Shapiro, and Daniel Cohen-Or.
\newblock Encoding in style: a stylegan encoder for image-to-image translation.
\newblock In {\em Proceedings of the IEEE/CVF conference on computer vision and pattern recognition}, pages 2287--2296, 2021.

\bibitem{riquelme2021scaling}
Carlos Riquelme, Joan Puigcerver, Basil Mustafa, Maxim Neumann, Rodolphe Jenatton, Andr{\'e} Susano~Pinto, Daniel Keysers, and Neil Houlsby.
\newblock Scaling vision with sparse mixture of experts.
\newblock {\em Advances in Neural Information Processing Systems}, 34:8583--8595, 2021.

\bibitem{sabour2017dynamic}
Sara Sabour, Nicholas Frosst, and Geoffrey~E Hinton.
\newblock Dynamic routing between capsules.
\newblock {\em Advances in neural information processing systems}, 30, 2017.

\bibitem{shen2020pdo}
Zhengyang Shen, Lingshen He, Zhouchen Lin, and Jinwen Ma.
\newblock Pdo-econvs: Partial differential operator based equivariant convolutions.
\newblock {\em arXiv preprint arXiv:2007.10408}, 2020.

\bibitem{sun2018igcv3}
Ke~Sun, Mingjie Li, Dong Liu, and Jingdong Wang.
\newblock Igcv3: Interleaved low-rank group convolutions for efficient deep neural networks.
\newblock {\em arXiv preprint arXiv:1806.00178}, 2018.

\bibitem{sur1990cross}
Mriganka Sur, Sarah~L Pallas, and Anna~W Roe.
\newblock Cross-modal plasticity in cortical development: differentiation and specification of sensory neocortex.
\newblock {\em Trends in neurosciences}, 13(6):227--233, 1990.

\bibitem{szegedy2015going}
Christian Szegedy, Wei Liu, Yangqing Jia, Pierre Sermanet, Scott Reed, Dragomir Anguelov, Dumitru Erhan, Vincent Vanhoucke, and Andrew Rabinovich.
\newblock Going deeper with convolutions.
\newblock In {\em Proceedings of the IEEE conference on computer vision and pattern recognition}, pages 1--9, 2015.

\bibitem{szegedy2016rethinking}
Christian Szegedy, Vincent Vanhoucke, Sergey Ioffe, Jon Shlens, and Zbigniew Wojna.
\newblock Rethinking the inception architecture for computer vision.
\newblock In {\em Proceedings of the IEEE conference on computer vision and pattern recognition}, pages 2818--2826, 2016.

\bibitem{takagi2023high}
Yu~Takagi and Shinji Nishimoto.
\newblock High-resolution image reconstruction with latent diffusion models from human brain activity.
\newblock In {\em Proceedings of the IEEE/CVF Conference on Computer Vision and Pattern Recognition}, pages 14453--14463, 2023.

\bibitem{toshev2014deeppose}
Alexander Toshev and Christian Szegedy.
\newblock Deeppose: Human pose estimation via deep neural networks.
\newblock In {\em Proceedings of the IEEE conference on computer vision and pattern recognition}, pages 1653--1660, 2014.

\bibitem{weiler2019general}
Maurice Weiler and Gabriele Cesa.
\newblock General e(2) - equivariant steerable cnns.
\newblock {\em arXiv preprint arXiv:1911.08251}, 2019.

\bibitem{weiler2018learning}
Maurice Weiler, Fred~A Hamprecht, and Martin Storath.
\newblock Learning steerable filters for rotation equivariant cnns.
\newblock In {\em Proceedings of the IEEE Conference on Computer Vision and Pattern Recognition}, pages 849--858, 2018.

\bibitem{Xie_2015_ICCV}
Saining Xie and Zhuowen Tu.
\newblock Holistically-nested edge detection.
\newblock In {\em Proceedings of the IEEE International Conference on Computer Vision (ICCV)}, December 2015.

\bibitem{xu2023magicanimate}
Zhongcong Xu, Jianfeng Zhang, Jun~Hao Liew, Hanshu Yan, Jia-Wei Liu, Chenxu Zhang, Jiashi Feng, and Mike~Zheng Shou.
\newblock Magicanimate: Temporally consistent human image animation using diffusion model.
\newblock {\em arXiv preprint arXiv:2311.16498}, 2023.

\bibitem{Zhang_2023_ICCV}
Lvmin Zhang, Anyi Rao, and Maneesh Agrawala.
\newblock Adding conditional control to text-to-image diffusion models.
\newblock In {\em Proceedings of the IEEE/CVF International Conference on Computer Vision (ICCV)}, pages 3836--3847, October 2023.

\bibitem{zhang2017interleaved}
Ting Zhang, Guo-Jun Qi, Bin Xiao, and Jingdong Wang.
\newblock Interleaved group convolutions.
\newblock In {\em Proceedings of the IEEE international conference on computer vision}, pages 4373--4382, 2017.

\bibitem{zhang2018shufflenet}
Xiangyu Zhang, Xinyu Zhou, Mengxiao Lin, and Jian Sun.
\newblock Shufflenet: An extremely efficient convolutional neural network for mobile devices.
\newblock In {\em Proceedings of the IEEE conference on computer vision and pattern recognition}, pages 6848--6856, 2018.

\bibitem{zhao2024uni}
Shihao Zhao, Dongdong Chen, Yen-Chun Chen, Jianmin Bao, Shaozhe Hao, Lu~Yuan, and Kwan-Yee~K Wong.
\newblock Uni-controlnet: All-in-one control to text-to-image diffusion models.
\newblock {\em Advances in Neural Information Processing Systems}, 36, 2024.

\bibitem{Zheng_2023_CVPR}
Guangcong Zheng, Xianpan Zhou, Xuewei Li, Zhongang Qi, Ying Shan, and Xi~Li.
\newblock Layoutdiffusion: Controllable diffusion model for layout-to-image generation.
\newblock In {\em Proceedings of the IEEE/CVF Conference on Computer Vision and Pattern Recognition (CVPR)}, pages 22490--22499, June 2023.

\bibitem{zhou2022mixture}
Yanqi Zhou, Tao Lei, Hanxiao Liu, Nan Du, Yanping Huang, Vincent Zhao, Andrew~M Dai, Quoc~V Le, James Laudon, et~al.
\newblock Mixture-of-experts with expert choice routing.
\newblock {\em Advances in Neural Information Processing Systems}, 35:7103--7114, 2022.

\bibitem{zhu2017unpaired}
Jun-Yan Zhu, Taesung Park, Phillip Isola, and Alexei~A Efros.
\newblock Unpaired image-to-image translation using cycle-consistent adversarial networks.
\newblock In {\em Proceedings of the IEEE international conference on computer vision}, pages 2223--2232, 2017.

\end{thebibliography}
\bibliographystyle{plain}

\newpage
\appendix
\beginsupplement
\section{Appendix / supplemental material}

\begin{figure}
  \centering
  \includegraphics[width=0.97\textwidth]{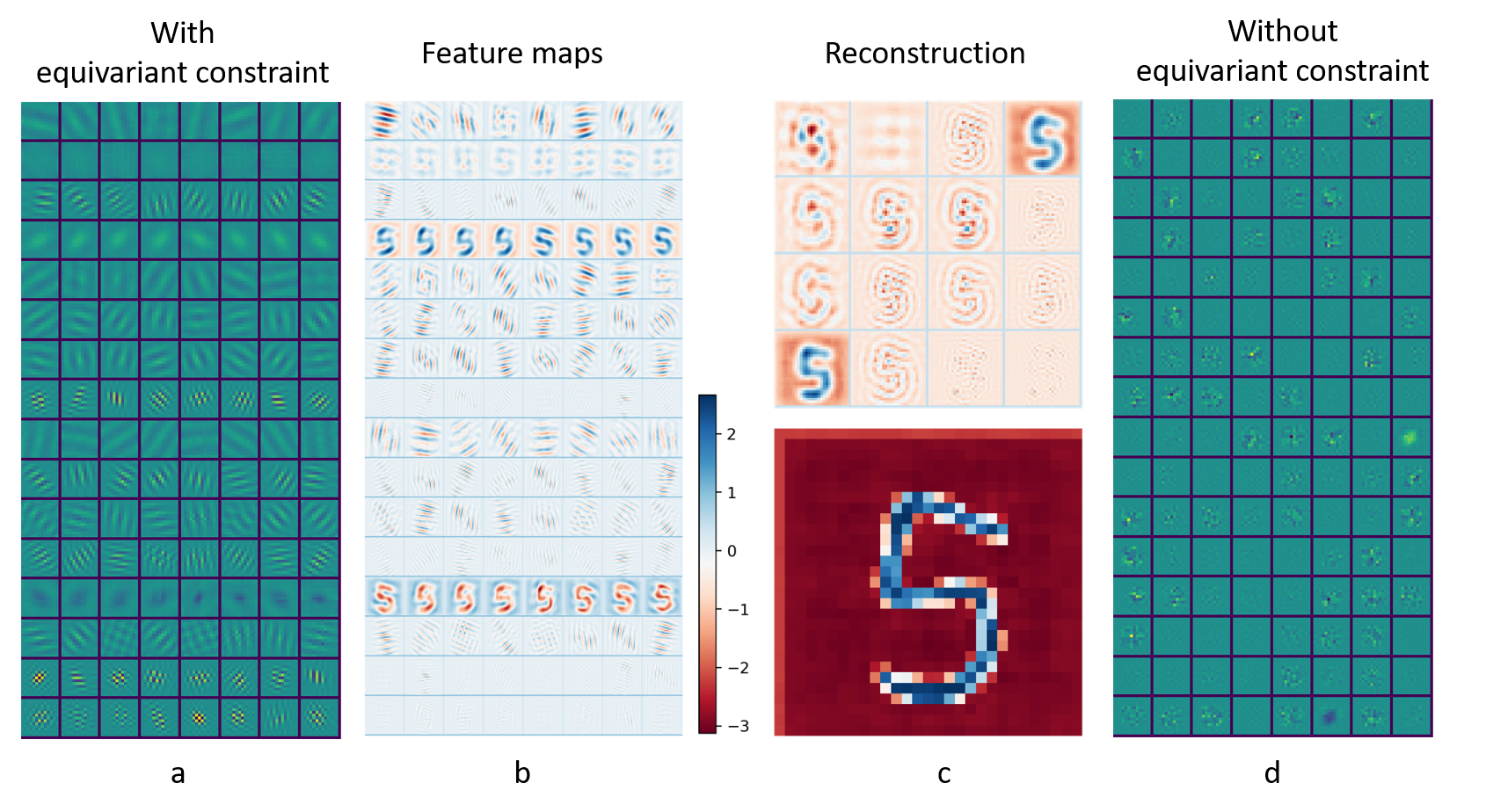}
  \caption{Ablation study on equivariance constraint.}\label{appdix_fig1_modulardiff}
\end{figure}

\begin{figure}
  \centering
  \includegraphics[width=\textwidth]{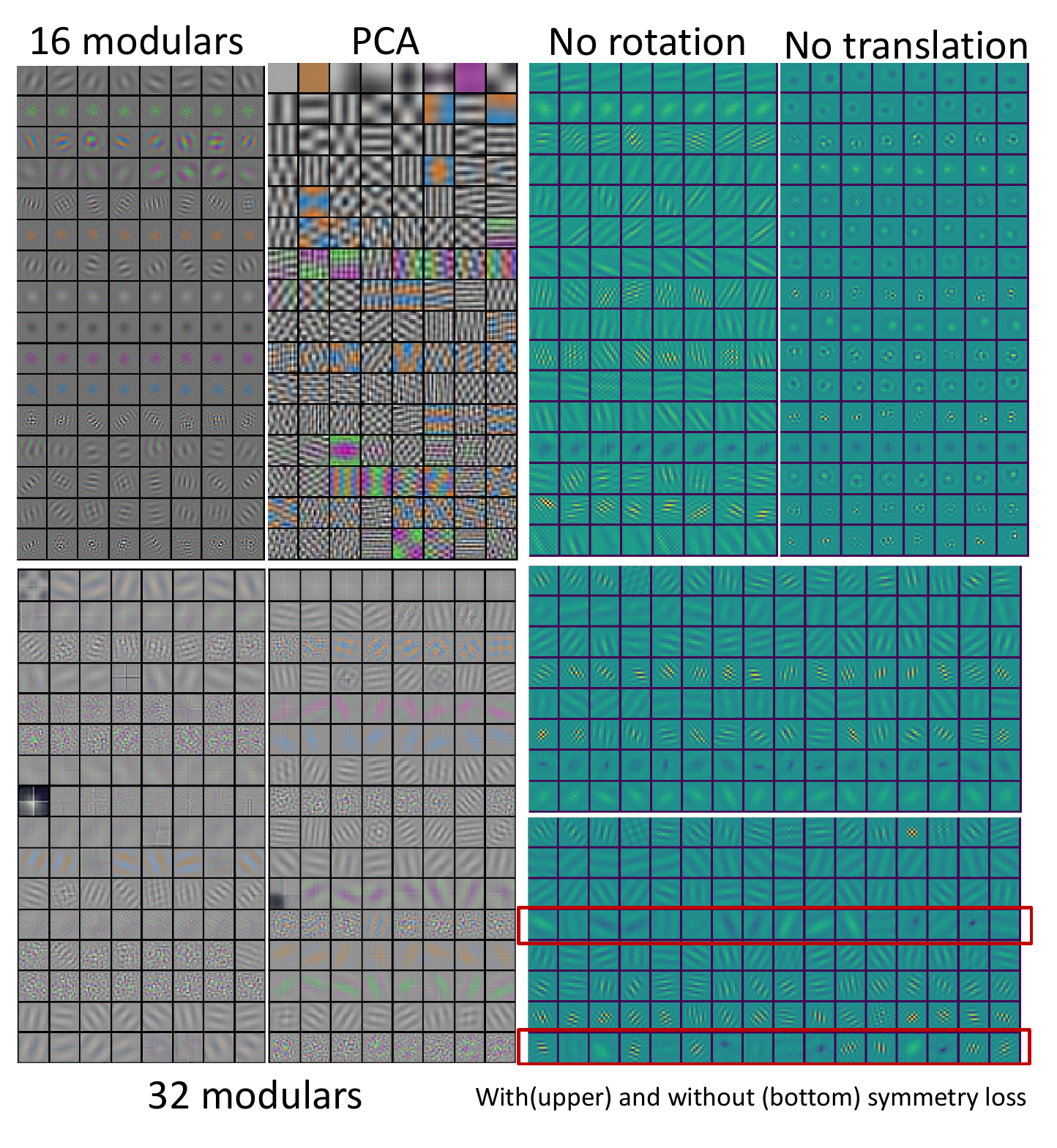}
  \caption{Ablation study on Modular Autoencoder.}\label{figr3}
\end{figure}

\begin{figure}
  \centering
  \includegraphics[width=0.97\textwidth]{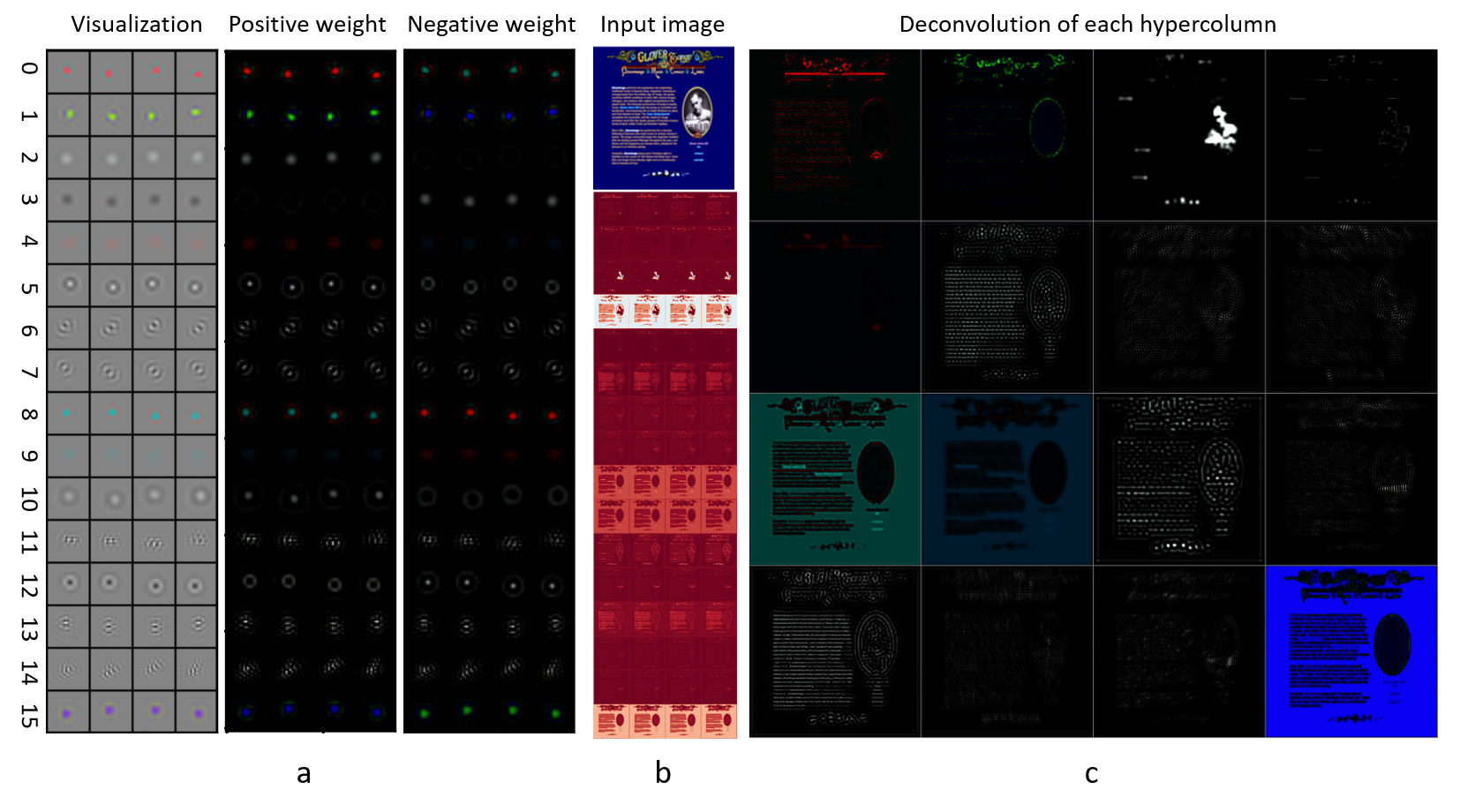}
  \caption{Color antagonism and center-surround receptive fields.}\label{appendix_modulardivide}
\end{figure}

\subsection{equivariance constraint}\label{appendix_equ_cons}
equivariance constraints are crucial for modular autoencoders. In this work, we designed simple equivariance constraints to validate the proposed modular autoencoder framework. We primarily employed two equivariance constraints: translational equivariance constraints and translational-rotational equivariance constraints. These constraints were applied to different module partitions. Since translational equivariance is encompassed by translational-rotational equivariance, modules corresponding to translational equivariance should be included within those corresponding to translational-rotational equivariance. When multiple translational modules are combined into a translational-rotational module, the learned features exhibit pronounced orientation selectivity, as illustrated in Figure \ref{appdix_fig1_modulardiff}a. Each convolutional kernel constitutes a translational equivariant module, and every 8 translational equivariant modules (a row) combine to form a translational-rotational equivariant module, termed a hypercolumn. Figure \ref{appendix_manifold}b demonstrates that using two convolutional kernels to constitute a translational equivariant module, and eight such modules to form a translational-rotational equivariant module, also results in learned orientation selectivity. When translational and translational-rotational modules coincide, as depicted in Figure \ref{appendix_modulardivide}a, each row represents both a translational equivariant module and a translational-rotational equivariant module. The learned features exhibit rotational symmetry. Regardless of the specific variations in the equivariant module settings, the autoencoder consistently produces clear functional specialization, demonstrating the effectiveness of the proposed equivariance constraints in achieving functional differentiation.

\subsection{Ablation Study on Equivariance Constraint and Modular Autoencoder}\label{appendix_equ_cons_ablation}

To validate the efficacy of equivariance constraints, we trained a modular autoencoder without utilizing these constraints under the same settings. Figure \ref{appdix_fig1_modulardiff}a depicts the learned features with equivariance constraints, while Figure \ref{appdix_fig1_modulardiff}b and c visualize the feature maps and reconstruction results of each module or hypercolumn individually. Notably, each module reconstructs images with a distinct focus on different spatial frequencies. When equivariance constraints are not employed, as shown in Figure \ref{appdix_fig1_modulardiff}d, the learned features fail to exhibit meaningful functional specialization or acquire hypercolumn-like orientation selectivity. This underscores the significance of our equivariance constraints in facilitating functional specialization and the emergence of hypercolumn-like characteristics.

By removing the symmetric loss (see Fig. \ref{figr3}), the autoencoder overall still works, but some modules exhibit unrelated sub-features within them. When we change the number of modules, the autoencoder still reliably forms feature differentiation. When removing the translation transform of the data, the learned features lose their direction selectivity. When removing the rotation transform, each module can only learn features with the same orientation. We also compare our modular features with PCA features. As shown in Fig \ref{figr3} we can see a clear difference compared to our approach, namely the lack of modular features in PCA. Theoretically, PCA, due to its lack of specialized modular features, struggles to flexibly manipulate different aspects of features, leading to a generative model that is almost a reconstruction of the original image and and thus lacks the ability to achieve widespread zero-shot generalization.

\subsection{Brain-like Characteristics}

\begin{figure}
  \centering
  \includegraphics[width=\textwidth]{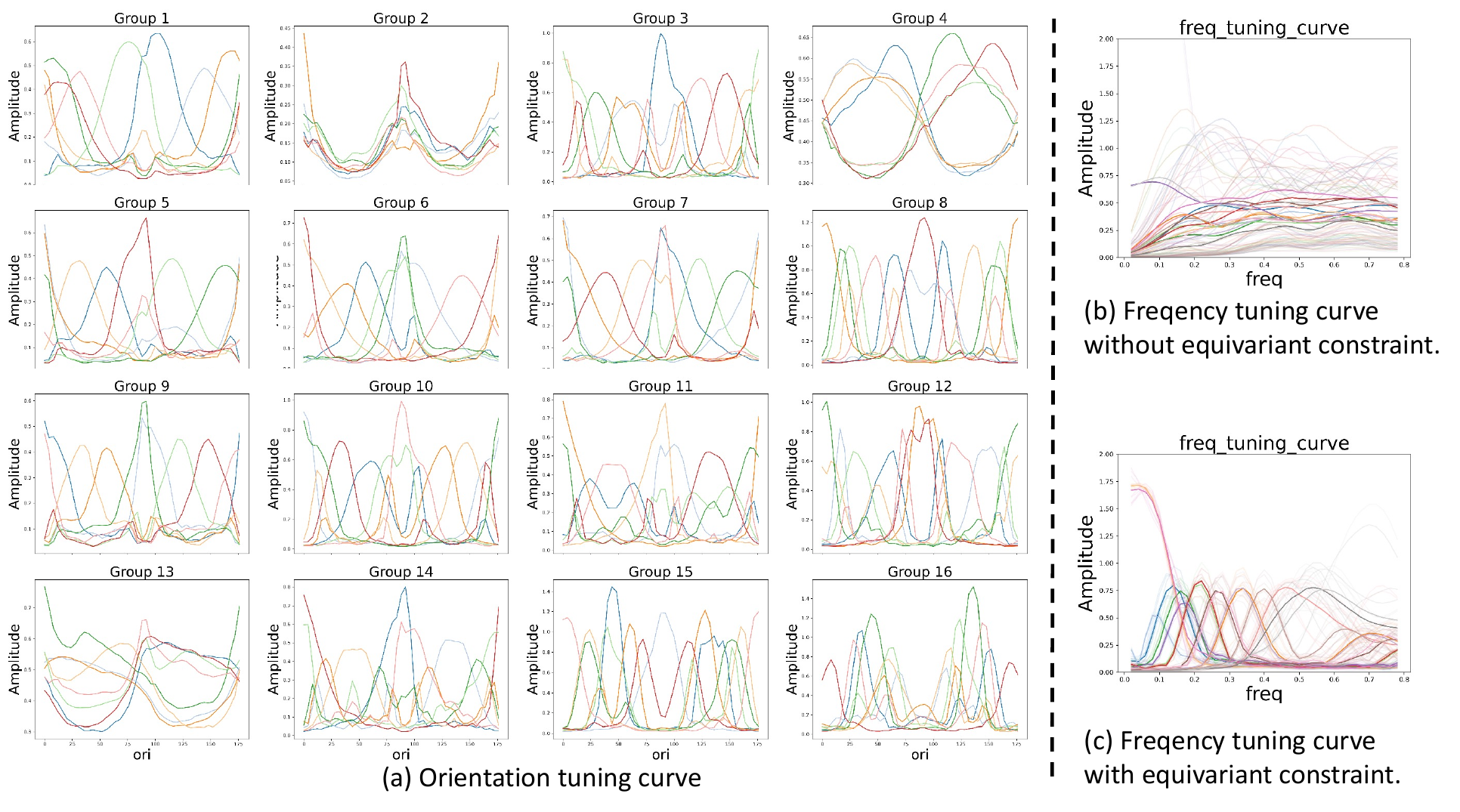}
  \caption{Orientation and frequency tuning curves.}\label{appendix_tuningcurve}
\end{figure}

\begin{figure}
  \centering
  \includegraphics[width=\textwidth]{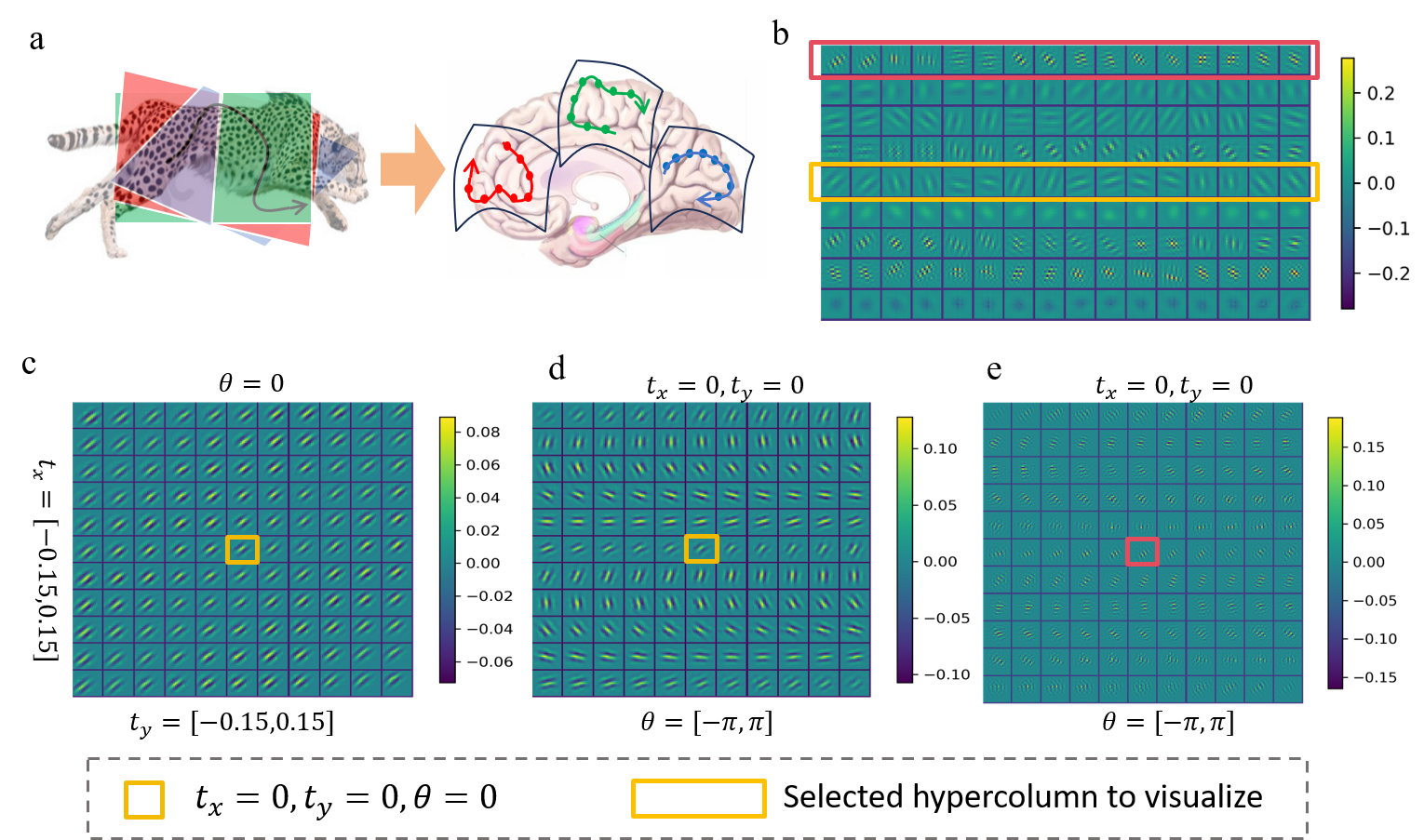}
  \caption{Close and complete submanifold for each hypercolumn.}\label{appendix_manifold}
\end{figure}

\subsubsection{Brain-like Tuning Curve}\label{appendix_tuning_curve}
To further analyze the functional specialization and brain-like characteristics of the modules, we tested the orientation and frequency tuning curves of each kernel using grating images of varying frequencies and orientations, as illustrated in Figure \ref{appendix_tuningcurve}. In Figure \ref{appendix_tuningcurve}a, each plot represents a module or hypercolumn, with each curve representing the orientation tuning curve of a kernel. Notably, most kernels spontaneously learned orientation selectivity, exhibiting a Gaussian-shaped tuning curve. Figure \ref{appendix_tuningcurve}b and \ref{appendix_tuningcurve}c depict the overall spatial frequency tuning curves of each module with and without equivariance constraints, respectively. Thin lines and thick lines represent the tuning curves of individual kernels and modules, respectively. In the absence of equivariance constraints, the learned features do not exhibit significant spatial frequency differentiation or clear frequency selectivity between modules. However, in the presence of equivariance constraints, the modules demonstrate highly pronounced frequency selectivity, characterized by distinct Gaussian-shaped tuning curves. Collectively, these modules effectively cover a relatively complete frequency space.

\subsubsection{Population Coding and Submanifold}\label{appendix_population_submanifold}
Each hypercolumn encompasses a complete orientation space, forming a self-contained and complete spatial representation, as visually demonstrated in Figure \ref{appendix_manifold}. Figure \ref{appendix_manifold}b depicts the experimental hypercolumn. By activating a single neuron within a hypercolumn and subsequently utilizing the prediction matrix $M^{(i)}(\delta)$ to predict and visualize the response pattern of the hypercolumn under various transformations, Figure \ref{appendix_manifold}c showcases the visualization of the predicted representation for translational reconstruction, revealing continuous spatial translations in the reconstructed images. Figure \ref{appendix_manifold}d presents the visualization of the predicted representation for rotational reconstruction, demonstrating continuous and complete orientation variations. These reconstructions are obtained through linear combinations of kernels within the same hypercolumn, echoing the phenomenon of population coding in biological systems. This suggests that each hypercolumn constitutes a closed and complete submanifold space. Figure \ref{appendix_manifold}e illustrates the orientation representation visualization of another hypercolumn, highlighting that different hypercolumns form distinct submanifold spaces. This formation of independent, closed, and complete submanifold spaces indicates that our equivariance constraints effectively enhance intra-modular correlations and inter-modular independence, thereby facilitating functional specialization among modules.

\subsubsection{Color Antagonism and Center-surround Receptive Fields}\label{appendix_color_centersurround}
As shown in Figure \ref{appendix_modulardivide}, when translational equivariant modules and translational-rotational equivariant modules coincide, rotationally symmetric features can be learned. These features exhibit a prominent center-surround antagonistic receptive field, as observed in hypercolumns 5, 10, and 12 in Figure \ref{appendix_modulardivide}a. This bears a resemblance to the on-center, off-center cells found in the biological primary visual cortex. For hypercolumns 1, 2, 8, and 15, distinct color selectivity emerges, accompanied by a prominent center-surround receptive field and the presence of color antagonism, a characteristic commonly observed in biological systems. These brain-like center-surround receptive fields and color antagonism are believed to enhance robustness to noise.

\subsection{More results for SCG}
\begin{figure}
    \centering
    \includegraphics[width=\textwidth]{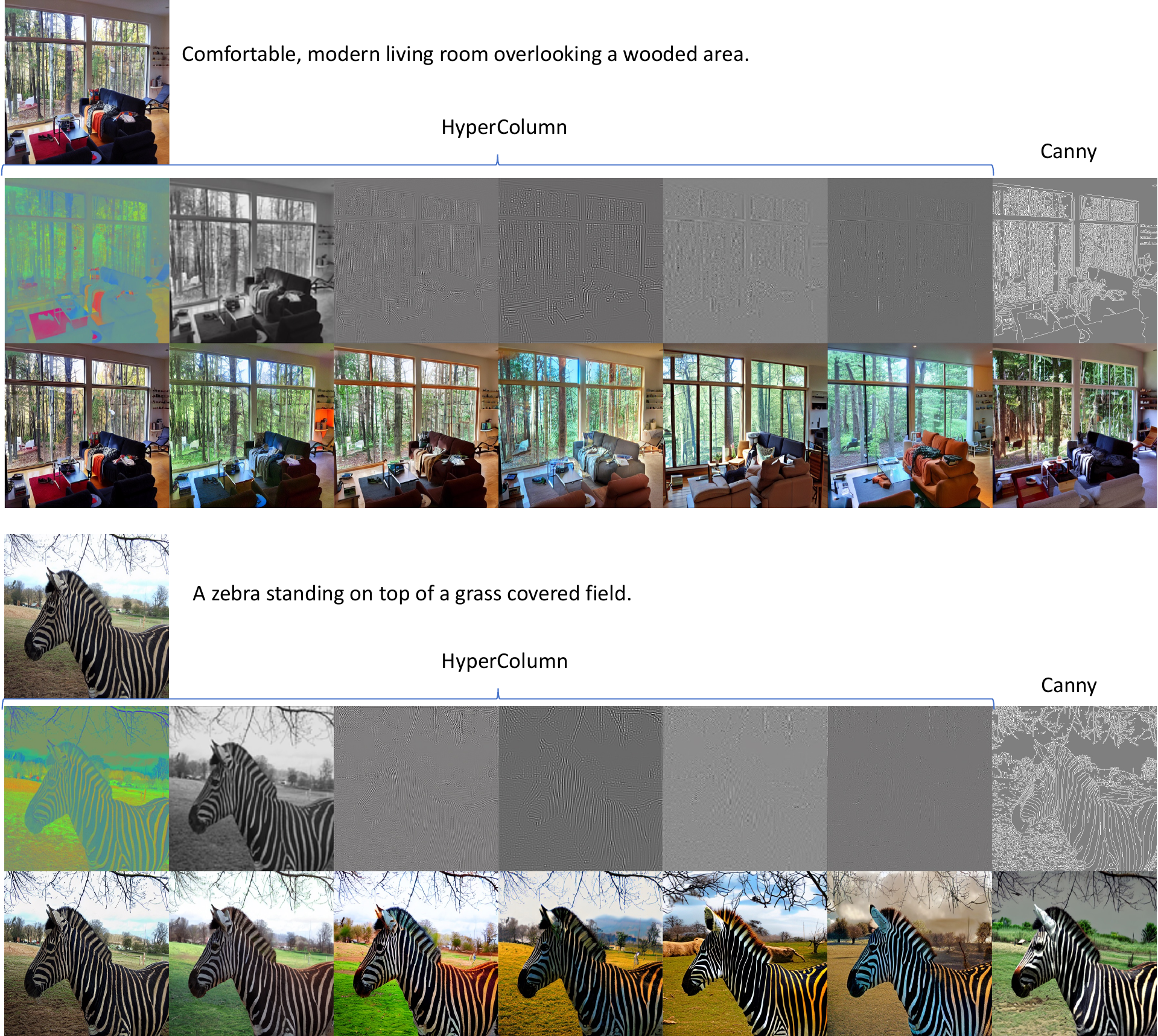}
    \caption{More generated images in COCO.}
    \label{appendix_cocoval}
\end{figure}

\begin{figure}
    \centering
    \includegraphics[width=\textwidth]{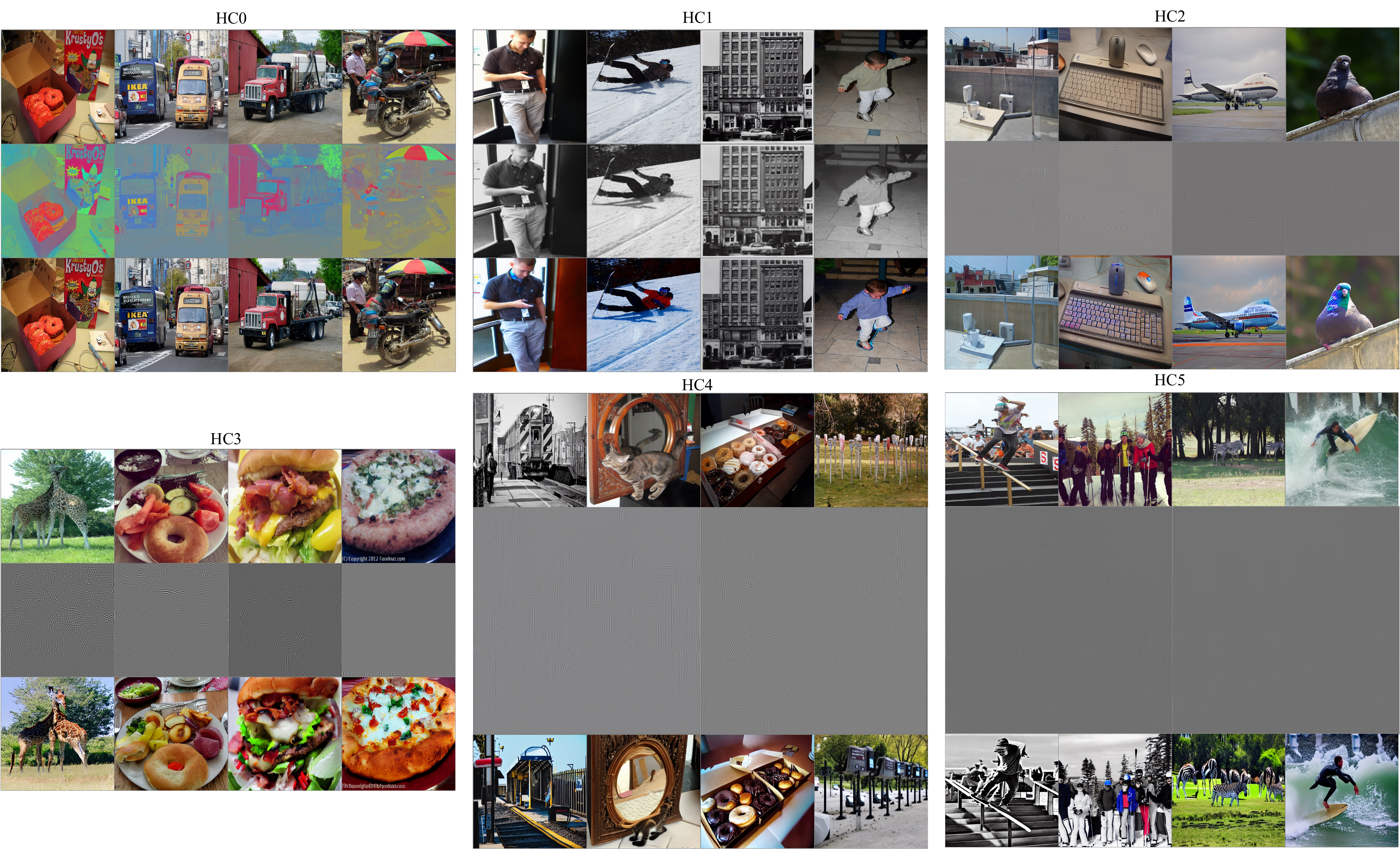}
    \caption{More generated images in COCO.}
    \label{appendix_cocotr}
\end{figure}

\begin{figure}
  \centering
  \includegraphics[width=\textwidth]{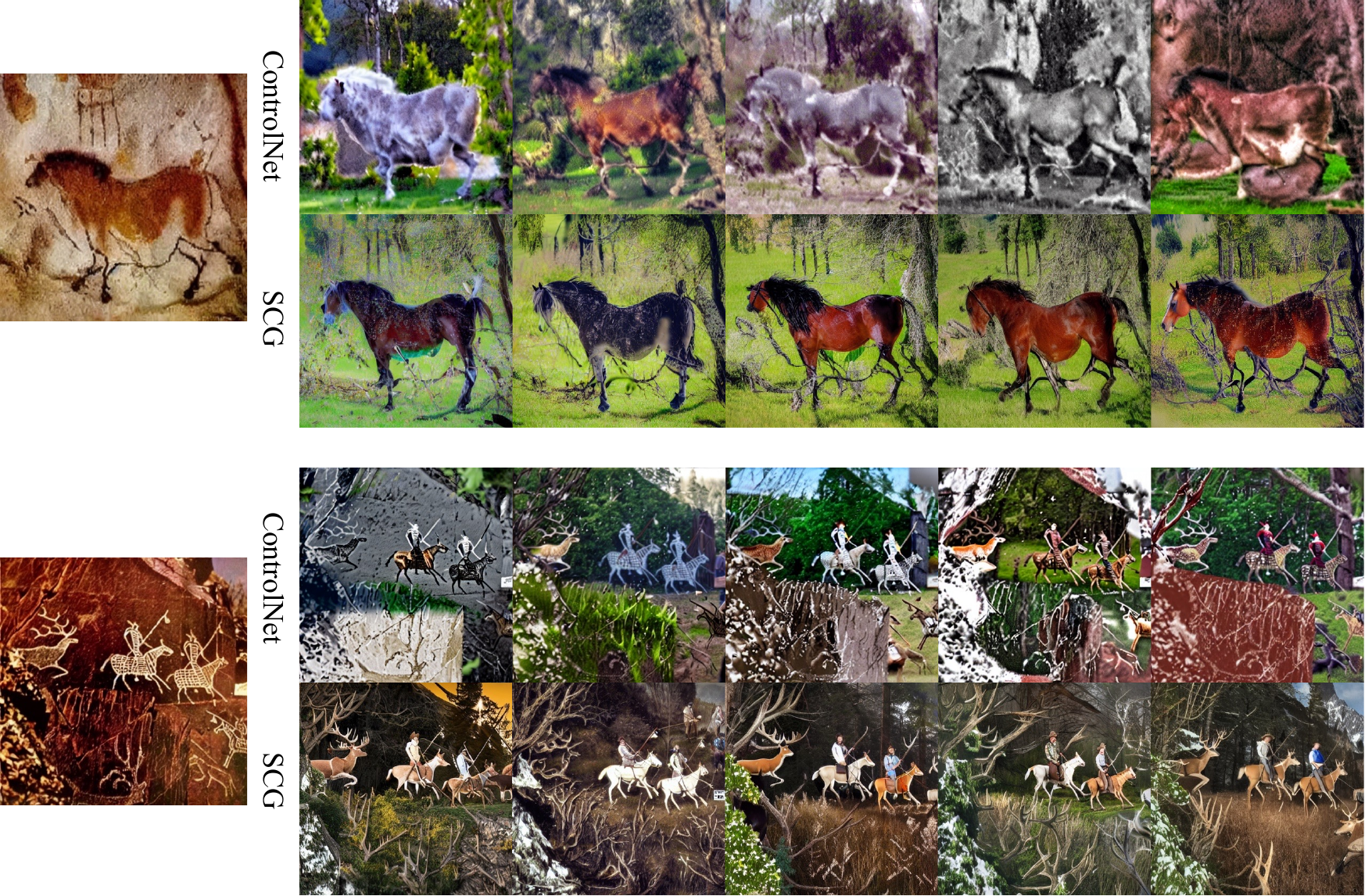}
  \caption{More results for ancient graffiti.}\label{appendix_bihua}
\end{figure}

\begin{figure}
  \centering
  \includegraphics[width=0.95\textwidth]{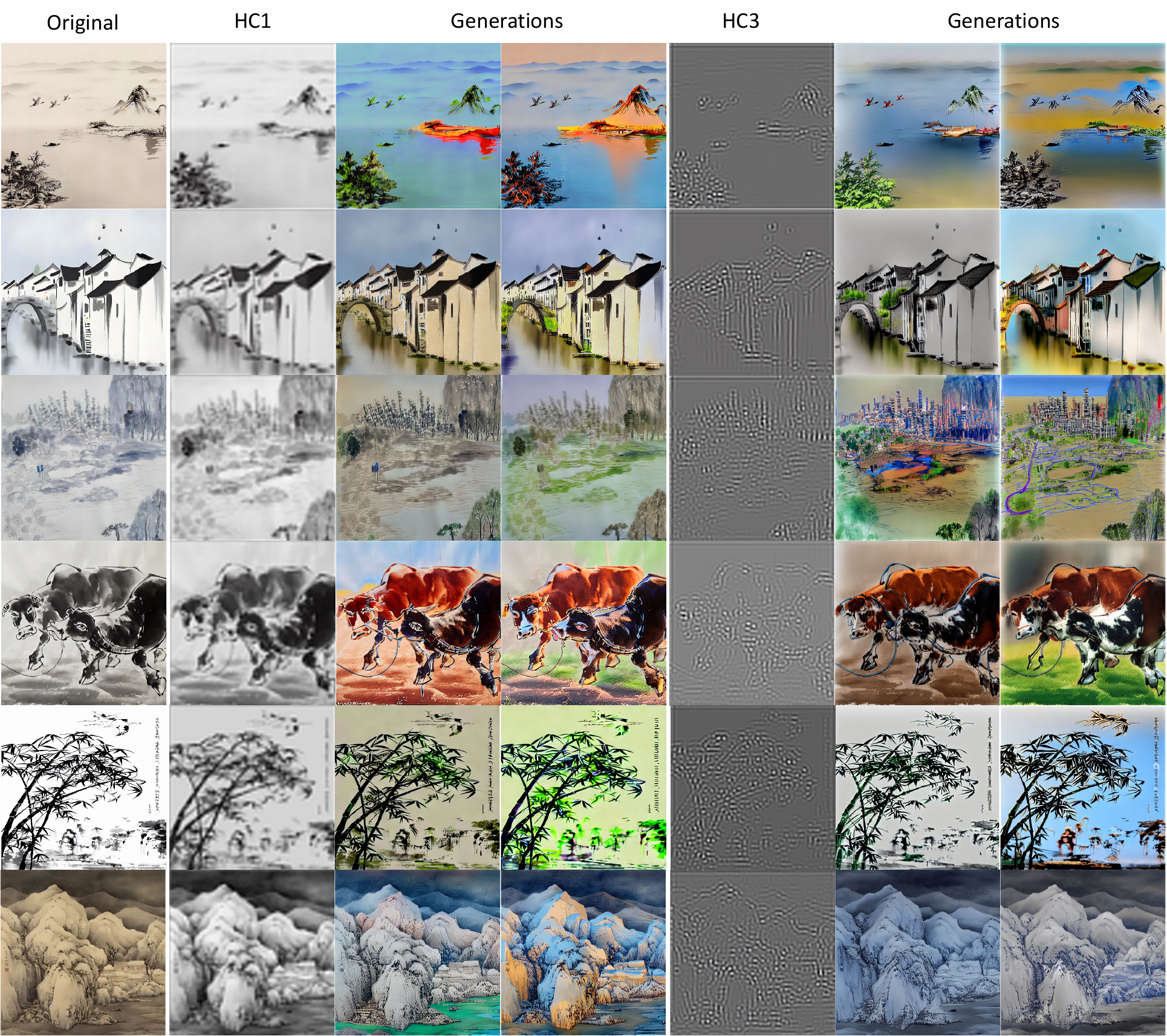}
  \caption{More results for wash and ink painting.}\label{appdix_fig1_shuimo}
\end{figure}

\begin{figure}
  \centering
  \includegraphics[width=\textwidth]{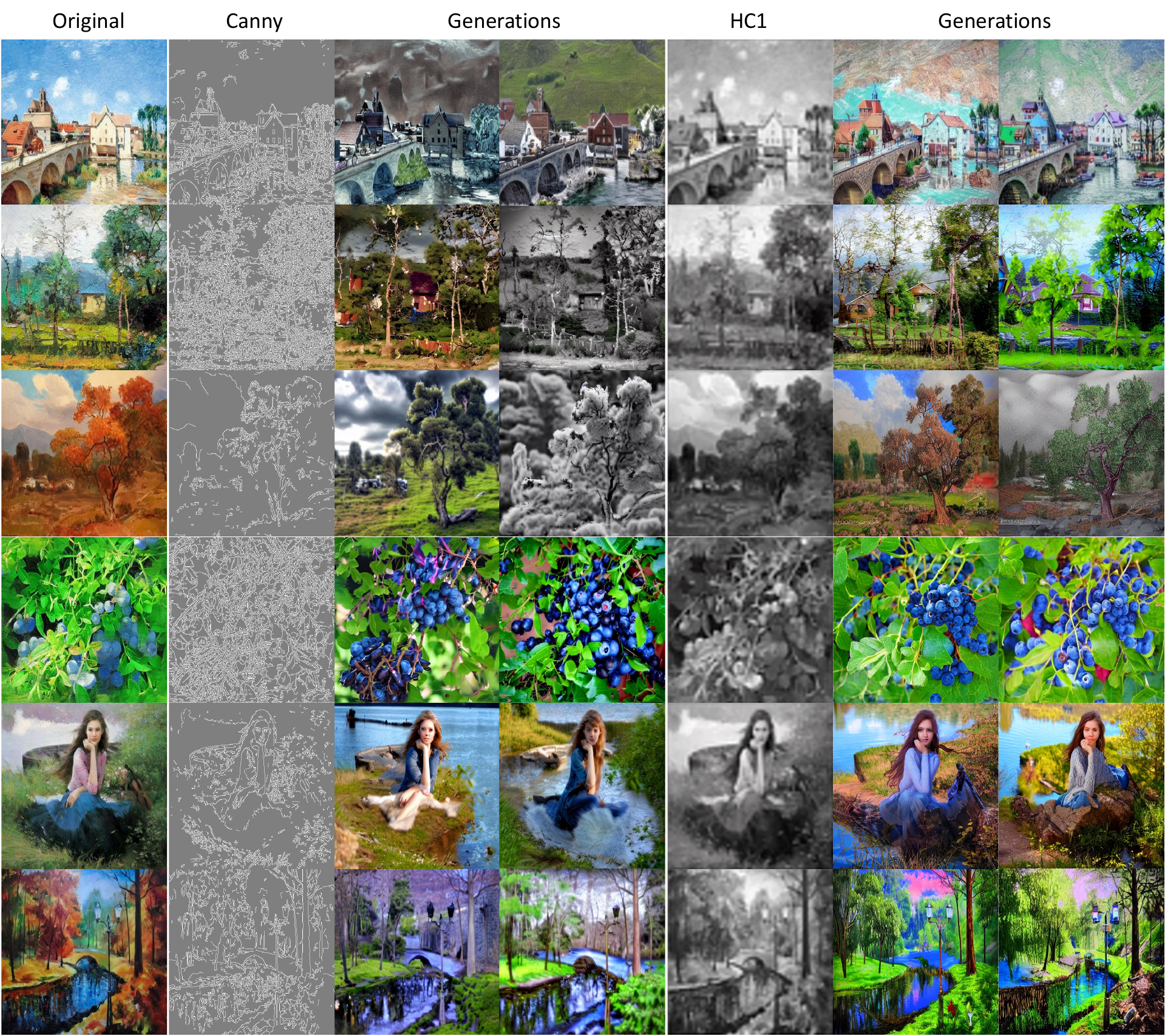}
  \caption{More results for oil painting.}\label{appdix_fig1_oil}
\end{figure}

\subsubsection{Experiment Details}\label{appendix_experiment_details}
During the self-supervised pattern completion stage, we utilize the differentiated modules from Figure \ref{appendix_modulardivide}a as feature extractors. We divide the 16 modules into six parts: modules 1, 2, 4, 8, 9, and 15 form HC0; modules 2 and 3 form HC1; modules 5 and 12 form HC2; module 10 forms HC3; modules 6 and 7 form HC4; and modules 11, 13, and 14 form HC5. Accordingly, HC0 comprises all modules sensitive to color, HC1 comprises modules sensitive to brightness, HC2 and HC3 comprise modules sensitive to edges, and HC4 and HC5 comprise modules sensitive to higher frequency features.

\subsubsection{Training Setup}\label{appendix_training_setup}
On the pattern completion stage, we employed ControlNet as our backbone network. We utilized SD1.5 as the pretrained diffusion model, which remained frozen during training. For the control conditioning branch, we extracted features from the selected modules and upsampled them to pixel space to serve as control images. We trained our model for 5 epochs with a batch size of 4 on the COCO dataset using an NVIDIA A100 GPU. For the ControlNet baseline reported in the paper, we followed the same training procedure. We used random translation and rotation augmentation within 0.3 times the image length and 360 degrees for training our modular autoencoder. The modular artoencoder is a single-layer or can be equivalently considered as a single-layer network in our implementation. We train it for 40000 steps with AdamW optimizer on a cosine anneal strategy with a start learning rate of 0.005.
\begin{figure}
  \centering
  \includegraphics[width=\textwidth]{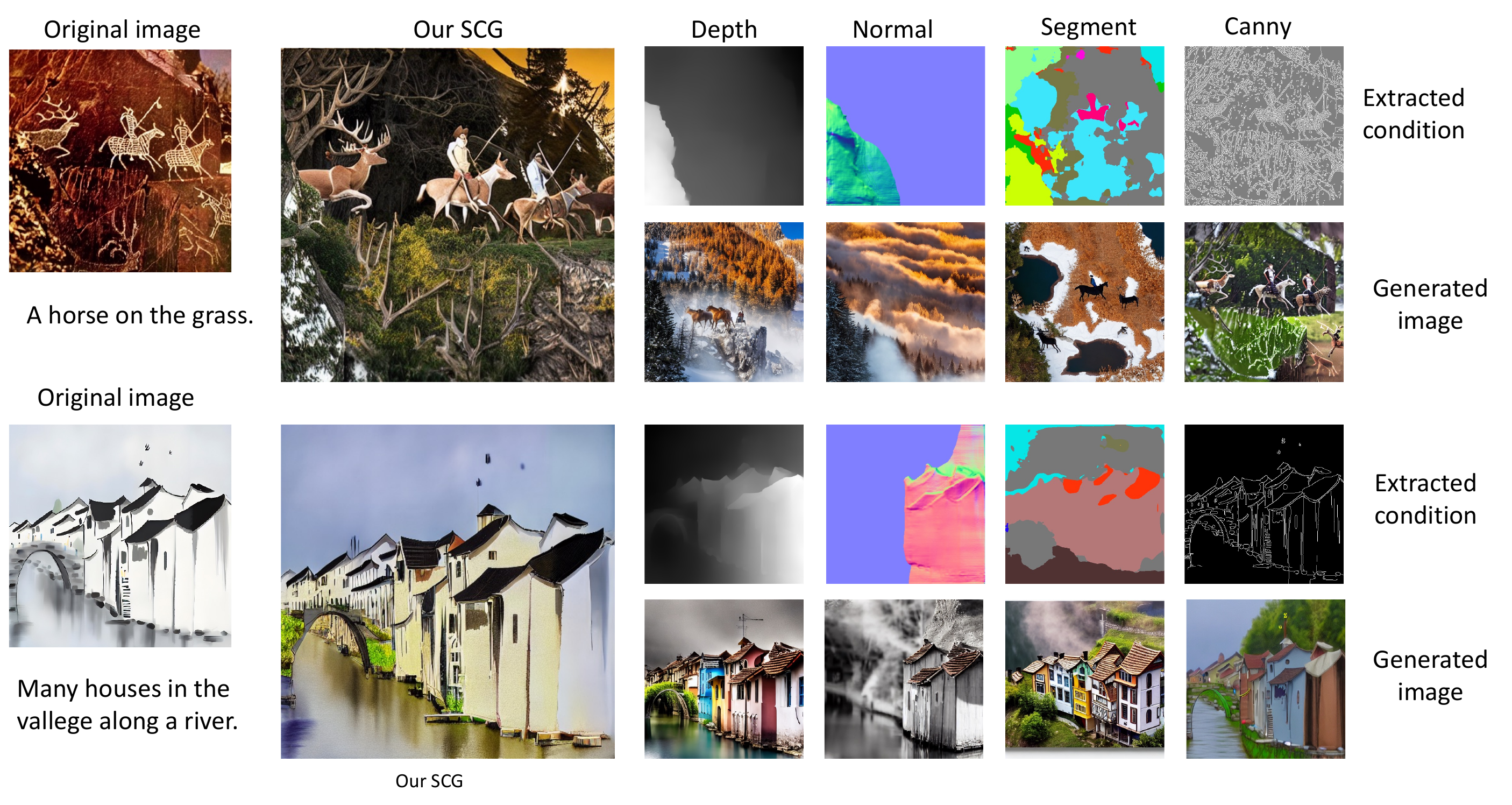}
  \caption{Comparisons under more conditions. Depth, Normal, Segment all extracted by pretrained models.}\label{figr1}
  
\end{figure}\begin{figure}
  \centering
  \includegraphics[width=\textwidth]{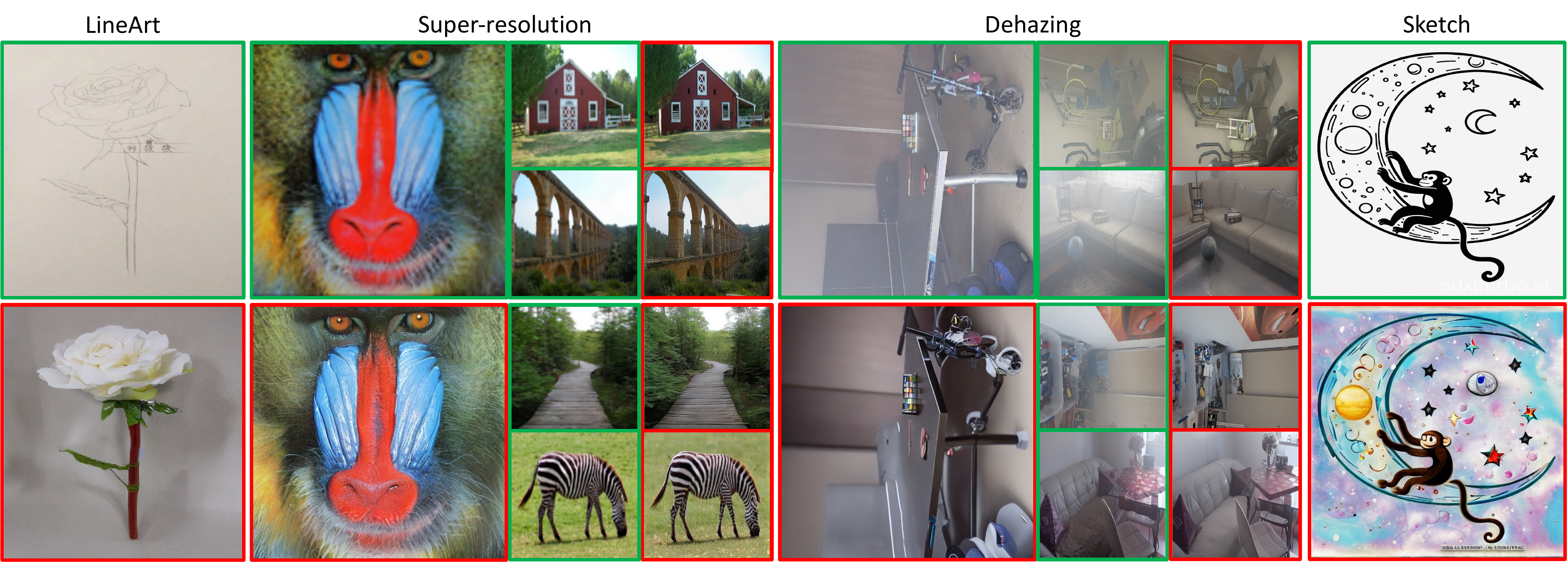}
  \caption{Zero-shot conditional generation of SCG on more tasks. With green box are conditions and with red box are generated images.}\label{figr2}
\end{figure}

\subsubsection{Datasets}
The MNIST database\footnote[1]{http://yann.lecun.com/exdb/mnist/} of handwritten digits has a training set of 60,000 examples, and a test set of 10,000 examples, where the digits have been size-normalized and centered in a fixed-size image.

ImageNet\footnote[2]{http://www.image-net.org/challenges/LSVRC/} Large Scale Visual Recognition Challenge (ILSVRC) contains over 1.2 million various size images of 1,000 classes for training and 50,000 images for validation. As the time and computing resources limits, we utilize 100 instead of the whole 1000 classes to train our modular autoencoder.

MS-COCO (Common Objects in Context) is a large-scale dataset for object detection, segmentation, captioning, and other computer vision tasks. It contains over 200,000 images with more than 1.5 million labeled objects. We use coco2017 train set including of 118K images to train our SCG.

\subsubsection{Subjective Evaluation}\label{appendix_subjective_evaluation}
We primarily conducted subjective evaluations on oil paintings and ancient graffiti. For oil paintings, we used six images and generated two pairs of images for each image using both ControlNet and SCG, resulting in a total of 12 image pairs. Participants were asked to rate which image in each pair had better fidelity and aesthetics. Finally, we statistics the mean winning rates for ControlNet and SCG. A total of 40 participants were recruited for this study, and all test images are shown in Figure \ref{appdix_fig1_oil}. For the ancient graffiti, a total of 37 participants were recruited for this study. And we statistics the winning rates for each pair generation. We use plattform https://www.wjx.cn/ to collect subjective evaluations. Participants were asked which image, out of the two, had better fidelity and aesthetics basing on the original image. 

\subsubsection{More Generations}\label{appendix_more_generation}
Figure \ref{appendix_cocoval} and Figure \ref{appendix_cocotr} showcase additional generation results on the COCO dataset using different hypercolumns as conditions. Figure \ref{appdix_fig1_shuimo} and \ref{appdix_fig1_oil} showcase additional generation results on the associative generation from wash and ink painting and oil painting. Figure \ref{appendix_bihua} shows more generation results for graffiti. It is evident that ControlNet-based generations are significantly impacted by irrelevant noise present in the ancient graffiti, leading to either blurry images or an abundance of unnatural details. In contrast, our proposed SCG method produces clearer images with more natural and realistic details, demonstrating superior robustness to noise.


\newpage
\section*{NeurIPS Paper Checklist}

\begin{enumerate}

\item {\bf Claims}
    \item[] Question: Do the main claims made in the abstract and introduction accurately reflect the paper's contributions and scope?
    \item[] Answer: \answerYes{} 
    \item[] Justification: See in section \ref{experiment}
    \item[] Guidelines:
    \begin{itemize}
        \item The answer NA means that the abstract and introduction do not include the claims made in the paper.
        \item The abstract and/or introduction should clearly state the claims made, including the contributions made in the paper and important assumptions and limitations. A No or NA answer to this question will not be perceived well by the reviewers. 
        \item The claims made should match theoretical and experimental results, and reflect how much the results can be expected to generalize to other settings. 
        \item It is fine to include aspirational goals as motivation as long as it is clear that these goals are not attained by the paper. 
    \end{itemize}

\item {\bf Limitations}
    \item[] Question: Does the paper discuss the limitations of the work performed by the authors?
    \item[] Answer: \answerYes{} 
    \item[] Justification: See in section \ref{limitation}
    \item[] Guidelines:
    \begin{itemize}
        \item The answer NA means that the paper has no limitation while the answer No means that the paper has limitations, but those are not discussed in the paper. 
        \item The authors are encouraged to create a separate "Limitations" section in their paper.
        \item The paper should point out any strong assumptions and how robust the results are to violations of these assumptions (e.g., independence assumptions, noiseless settings, model well-specification, asymptotic approximations only holding locally). The authors should reflect on how these assumptions might be violated in practice and what the implications would be.
        \item The authors should reflect on the scope of the claims made, e.g., if the approach was only tested on a few datasets or with a few runs. In general, empirical results often depend on implicit assumptions, which should be articulated.
        \item The authors should reflect on the factors that influence the performance of the approach. For example, a facial recognition algorithm may perform poorly when image resolution is low or images are taken in low lighting. Or a speech-to-text system might not be used reliably to provide closed captions for online lectures because it fails to handle technical jargon.
        \item The authors should discuss the computational efficiency of the proposed algorithms and how they scale with dataset size.
        \item If applicable, the authors should discuss possible limitations of their approach to address problems of privacy and fairness.
        \item While the authors might fear that complete honesty about limitations might be used by reviewers as grounds for rejection, a worse outcome might be that reviewers discover limitations that aren't acknowledged in the paper. The authors should use their best judgment and recognize that individual actions in favor of transparency play an important role in developing norms that preserve the integrity of the community. Reviewers will be specifically instructed to not penalize honesty concerning limitations.
    \end{itemize}

\item {\bf Theory Assumptions and Proofs}
    \item[] Question: For each theoretical result, does the paper provide the full set of assumptions and a complete (and correct) proof?
    \item[] Answer: \answerNo{} 
    \item[] Justification: We have no theory to proof.
    \item[] Guidelines:
    \begin{itemize}
        \item The answer NA means that the paper does not include theoretical results. 
        \item All the theorems, formulas, and proofs in the paper should be numbered and cross-referenced.
        \item All assumptions should be clearly stated or referenced in the statement of any theorems.
        \item The proofs can either appear in the main paper or the supplemental material, but if they appear in the supplemental material, the authors are encouraged to provide a short proof sketch to provide intuition. 
        \item Inversely, any informal proof provided in the core of the paper should be complemented by formal proofs provided in appendix or supplemental material.
        \item Theorems and Lemmas that the proof relies upon should be properly referenced. 
    \end{itemize}

    \item {\bf Experimental Result Reproducibility}
    \item[] Question: Does the paper fully disclose all the information needed to reproduce the main experimental results of the paper to the extent that it affects the main claims and/or conclusions of the paper (regardless of whether the code and data are provided or not)?
    \item[] Answer: \answerYes{} 
    \item[] Justification: See in section \ref{experiment}
    \item[] Guidelines:
    \begin{itemize}
        \item The answer NA means that the paper does not include experiments.
        \item If the paper includes experiments, a No answer to this question will not be perceived well by the reviewers: Making the paper reproducible is important, regardless of whether the code and data are provided or not.
        \item If the contribution is a dataset and/or model, the authors should describe the steps taken to make their results reproducible or verifiable. 
        \item Depending on the contribution, reproducibility can be accomplished in various ways. For example, if the contribution is a novel architecture, describing the architecture fully might suffice, or if the contribution is a specific model and empirical evaluation, it may be necessary to either make it possible for others to replicate the model with the same dataset, or provide access to the model. In general. releasing code and data is often one good way to accomplish this, but reproducibility can also be provided via detailed instructions for how to replicate the results, access to a hosted model (e.g., in the case of a large language model), releasing of a model checkpoint, or other means that are appropriate to the research performed.
        \item While NeurIPS does not require releasing code, the conference does require all submissions to provide some reasonable avenue for reproducibility, which may depend on the nature of the contribution. For example
        \begin{enumerate}
            \item If the contribution is primarily a new algorithm, the paper should make it clear how to reproduce that algorithm.
            \item If the contribution is primarily a new model architecture, the paper should describe the architecture clearly and fully.
            \item If the contribution is a new model (e.g., a large language model), then there should either be a way to access this model for reproducing the results or a way to reproduce the model (e.g., with an open-source dataset or instructions for how to construct the dataset).
            \item We recognize that reproducibility may be tricky in some cases, in which case authors are welcome to describe the particular way they provide for reproducibility. In the case of closed-source models, it may be that access to the model is limited in some way (e.g., to registered users), but it should be possible for other researchers to have some path to reproducing or verifying the results.
        \end{enumerate}
    \end{itemize}

\item {\bf Open access to data and code}
    \item[] Question: Does the paper provide open access to the data and code, with sufficient instructions to faithfully reproduce the main experimental results, as described in supplemental material?
    \item[] Answer: \answerYes{} 
    \item[] Justification: Code is submit in supplemental material.
    \item[] Guidelines:
    \begin{itemize}
        \item The answer NA means that paper does not include experiments requiring code.
        \item Please see the NeurIPS code and data submission guidelines (\url{https://nips.cc/public/guides/CodeSubmissionPolicy}) for more details.
        \item While we encourage the release of code and data, we understand that this might not be possible, so “No” is an acceptable answer. Papers cannot be rejected simply for not including code, unless this is central to the contribution (e.g., for a new open-source benchmark).
        \item The instructions should contain the exact command and environment needed to run to reproduce the results. See the NeurIPS code and data submission guidelines (\url{https://nips.cc/public/guides/CodeSubmissionPolicy}) for more details.
        \item The authors should provide instructions on data access and preparation, including how to access the raw data, preprocessed data, intermediate data, and generated data, etc.
        \item The authors should provide scripts to reproduce all experimental results for the new proposed method and baselines. If only a subset of experiments are reproducible, they should state which ones are omitted from the script and why.
        \item At submission time, to preserve anonymity, the authors should release anonymized versions (if applicable).
        \item Providing as much information as possible in supplemental material (appended to the paper) is recommended, but including URLs to data and code is permitted.
    \end{itemize}

\item {\bf Experimental Setting/Details}
    \item[] Question: Does the paper specify all the training and test details (e.g., data splits, hyperparameters, how they were chosen, type of optimizer, etc.) necessary to understand the results?
    \item[] Answer: \answerYes{} 
    \item[] Justification: See in section \ref{experiment} and supplemental material.
    \item[] Guidelines:
    \begin{itemize}
        \item The answer NA means that the paper does not include experiments.
        \item The experimental setting should be presented in the core of the paper to a level of detail that is necessary to appreciate the results and make sense of them.
        \item The full details can be provided either with the code, in appendix, or as supplemental material.
    \end{itemize}

\item {\bf Experiment Statistical Significance}
    \item[] Question: Does the paper report error bars suitably and correctly defined or other appropriate information about the statistical significance of the experiments?
    \item[] Answer: \answerNA{} 
    \item[] Justification: Statistical significance is not applicable for image controllable generation.
    \item[] Guidelines:
    \begin{itemize}
        \item The answer NA means that the paper does not include experiments.
        \item The authors should answer "Yes" if the results are accompanied by error bars, confidence intervals, or statistical significance tests, at least for the experiments that support the main claims of the paper.
        \item The factors of variability that the error bars are capturing should be clearly stated (for example, train/test split, initialization, random drawing of some parameter, or overall run with given experimental conditions).
        \item The method for calculating the error bars should be explained (closed form formula, call to a library function, bootstrap, etc.)
        \item The assumptions made should be given (e.g., Normally distributed errors).
        \item It should be clear whether the error bar is the standard deviation or the standard error of the mean.
        \item It is OK to report 1-sigma error bars, but one should state it. The authors should preferably report a 2-sigma error bar than state that they have a 96\% CI, if the hypothesis of Normality of errors is not verified.
        \item For asymmetric distributions, the authors should be careful not to show in tables or figures symmetric error bars that would yield results that are out of range (e.g. negative error rates).
        \item If error bars are reported in tables or plots, The authors should explain in the text how they were calculated and reference the corresponding figures or tables in the text.
    \end{itemize}

\item {\bf Experiments Compute Resources}
    \item[] Question: For each experiment, does the paper provide sufficient information on the computer resources (type of compute workers, memory, time of execution) needed to reproduce the experiments?
    \item[] Answer: \answerYes{} 
    \item[] Justification: See in appendix \ref{appendix_training_setup}
    \item[] Guidelines:
    \begin{itemize}
        \item The answer NA means that the paper does not include experiments.
        \item The paper should indicate the type of compute workers CPU or GPU, internal cluster, or cloud provider, including relevant memory and storage.
        \item The paper should provide the amount of compute required for each of the individual experimental runs as well as estimate the total compute. 
        \item The paper should disclose whether the full research project required more compute than the experiments reported in the paper (e.g., preliminary or failed experiments that didn't make it into the paper). 
    \end{itemize}
    
\item {\bf Code Of Ethics}
    \item[] Question: Does the research conducted in the paper conform, in every respect, with the NeurIPS Code of Ethics \url{https://neurips.cc/public/EthicsGuidelines}?
    \item[] Answer: \answerYes{} 
    \item[] Justification: We conducted the research in the paper conform, in every respect, with the NeurIPS Code of Ethics.
    \item[] Guidelines:
    \begin{itemize}
        \item The answer NA means that the authors have not reviewed the NeurIPS Code of Ethics.
        \item If the authors answer No, they should explain the special circumstances that require a deviation from the Code of Ethics.
        \item The authors should make sure to preserve anonymity (e.g., if there is a special consideration due to laws or regulations in their jurisdiction).
    \end{itemize}

\item {\bf Broader Impacts}
    \item[] Question: Does the paper discuss both potential positive societal impacts and negative societal impacts of the work performed?
    \item[] Answer: \answerYes{} 
    \item[] Justification: See in section \ref{limitation}
    \item[] Guidelines:
    \begin{itemize}
        \item The answer NA means that there is no societal impact of the work performed.
        \item If the authors answer NA or No, they should explain why their work has no societal impact or why the paper does not address societal impact.
        \item Examples of negative societal impacts include potential malicious or unintended uses (e.g., disinformation, generating fake profiles, surveillance), fairness considerations (e.g., deployment of technologies that could make decisions that unfairly impact specific groups), privacy considerations, and security considerations.
        \item The conference expects that many papers will be foundational research and not tied to particular applications, let alone deployments. However, if there is a direct path to any negative applications, the authors should point it out. For example, it is legitimate to point out that an improvement in the quality of generative models could be used to generate deepfakes for disinformation. On the other hand, it is not needed to point out that a generic algorithm for optimizing neural networks could enable people to train models that generate Deepfakes faster.
        \item The authors should consider possible harms that could arise when the technology is being used as intended and functioning correctly, harms that could arise when the technology is being used as intended but gives incorrect results, and harms following from (intentional or unintentional) misuse of the technology.
        \item If there are negative societal impacts, the authors could also discuss possible mitigation strategies (e.g., gated release of models, providing defenses in addition to attacks, mechanisms for monitoring misuse, mechanisms to monitor how a system learns from feedback over time, improving the efficiency and accessibility of ML).
    \end{itemize}
    
\item {\bf Safeguards}
    \item[] Question: Does the paper describe safeguards that have been put in place for responsible release of data or models that have a high risk for misuse (e.g., pretrained language models, image generators, or scraped datasets)?
    \item[] Answer: \answerYes{} 
    \item[] Justification: We use exist excellent ControlNet and their safeguards. Our model will not bring more risk than ControlNet.
    \item[] Guidelines:
    \begin{itemize}
        \item The answer NA means that the paper poses no such risks.
        \item Released models that have a high risk for misuse or dual-use should be released with necessary safeguards to allow for controlled use of the model, for example by requiring that users adhere to usage guidelines or restrictions to access the model or implementing safety filters. 
        \item Datasets that have been scraped from the Internet could pose safety risks. The authors should describe how they avoided releasing unsafe images.
        \item We recognize that providing effective safeguards is challenging, and many papers do not require this, but we encourage authors to take this into account and make a best faith effort.
    \end{itemize}

\item {\bf Licenses for existing assets}
    \item[] Question: Are the creators or original owners of assets (e.g., code, data, models), used in the paper, properly credited and are the license and terms of use explicitly mentioned and properly respected?
    \item[] Answer: \answerYes{} 
    \item[] Justification: We use CC BY-NC 4.0 license.
    \item[] Guidelines:
    \begin{itemize}
        \item The answer NA means that the paper does not use existing assets.
        \item The authors should cite the original paper that produced the code package or dataset.
        \item The authors should state which version of the asset is used and, if possible, include a URL.
        \item The name of the license (e.g., CC-BY 4.0) should be included for each asset.
        \item For scraped data from a particular source (e.g., website), the copyright and terms of service of that source should be provided.
        \item If assets are released, the license, copyright information, and terms of use in the package should be provided. For popular datasets, \url{paperswithcode.com/datasets} has curated licenses for some datasets. Their licensing guide can help determine the license of a dataset.
        \item For existing datasets that are re-packaged, both the original license and the license of the derived asset (if it has changed) should be provided.
        \item If this information is not available online, the authors are encouraged to reach out to the asset's creators.
    \end{itemize}

\item {\bf New Assets}
    \item[] Question: Are new assets introduced in the paper well documented and is the documentation provided alongside the assets?
    \item[] Answer: \answerYes{} 
    \item[] Justification: We release new code based on ControlNet. The code is include an anonymized zip file.
    \item[] Guidelines:
    \begin{itemize}
        \item The answer NA means that the paper does not release new assets.
        \item Researchers should communicate the details of the dataset/code/model as part of their submissions via structured templates. This includes details about training, license, limitations, etc. 
        \item The paper should discuss whether and how consent was obtained from people whose asset is used.
        \item At submission time, remember to anonymize your assets (if applicable). You can either create an anonymized URL or include an anonymized zip file.
    \end{itemize}

\item {\bf Crowdsourcing and Research with Human Subjects}
    \item[] Question: For crowdsourcing experiments and research with human subjects, does the paper include the full text of instructions given to participants and screenshots, if applicable, as well as details about compensation (if any)? 
    \item[] Answer: \answerNA{} 
    \item[] Justification: We conducted a brief, unpaid survey on the Internet social App lasting less than 5 minutes for each survey. No crowdsourcing nor research with human subjects.
    \item[] Guidelines:
    \begin{itemize}
        \item The answer NA means that the paper does not involve crowdsourcing nor research with human subjects.
        \item Including this information in the supplemental material is fine, but if the main contribution of the paper involves human subjects, then as much detail as possible should be included in the main paper. 
        \item According to the NeurIPS Code of Ethics, workers involved in data collection, curation, or other labor should be paid at least the minimum wage in the country of the data collector. 
    \end{itemize}

\item {\bf Institutional Review Board (IRB) Approvals or Equivalent for Research with Human Subjects}
    \item[] Question: Does the paper describe potential risks incurred by study participants, whether such risks were disclosed to the subjects, and whether Institutional Review Board (IRB) approvals (or an equivalent approval/review based on the requirements of your country or institution) were obtained?
    \item[] Answer: \answerNA{} 
    \item[] Justification: Not applicable.
    \item[] Guidelines:
    \begin{itemize}
        \item The answer NA means that the paper does not involve crowdsourcing nor research with human subjects.
        \item Depending on the country in which research is conducted, IRB approval (or equivalent) may be required for any human subjects research. If you obtained IRB approval, you should clearly state this in the paper. 
        \item We recognize that the procedures for this may vary significantly between institutions and locations, and we expect authors to adhere to the NeurIPS Code of Ethics and the guidelines for their institution. 
        \item For initial submissions, do not include any information that would break anonymity (if applicable), such as the institution conducting the review.
    \end{itemize}

\end{enumerate}

\end{document}